\documentclass{article}

\usepackage[preprint]{neurips_2024}

\usepackage[utf8]{inputenc} 
\usepackage[T1]{fontenc}    
\usepackage{hyperref}       
\usepackage{url}            
\usepackage{booktabs}       
\usepackage{amsfonts}       
\usepackage{nicefrac}       
\usepackage{microtype}      
\usepackage{xcolor}         
\usepackage{graphicx}
\usepackage{subcaption}
\usepackage{titling} 
\usepackage[normalem]{ulem}  

\usepackage{amsmath}      
\usepackage{amssymb}      
\usepackage{amsthm}       
\usepackage{mathtools}    
\usepackage{bm}           

\theoremstyle{definition}
\newtheorem{definition}{Definition}[section]
\newtheorem{problem}{Problem}[section]

\theoremstyle{plain}
\newtheorem{theorem}{Theorem}[section]

\newtheorem{proposition}[theorem]{Proposition}

\theoremstyle{remark}
\newtheorem{remark}{Remark}[section]

\usepackage{booktabs}     
\usepackage{array}        
\usepackage{multirow}     
\usepackage{graphicx}     
\usepackage{subcaption}   
\usepackage[most]{tcolorbox}
\usepackage{enumitem}

\usepackage{enumitem}     
\setlist[itemize]{nosep, left=0pt}  
\setlist[enumerate]{nosep, left=0pt}  

\usepackage{algorithm}
\usepackage{algorithmic}
\usepackage{amssymb}  
\usepackage{pifont}   
\newcommand{\cmark}{\ding{51}}  
\newcommand{\xmark}{\ding{55}}  

\usepackage{natbib}       

\newcommand{\Cmas}{C_{\textsc{mas}}}
\newcommand{\CSAS}{C_{\textsc{SAS}}}
\newcommand{\Cselect}{C_{\text{select}}}
\newcommand{\Cexec}{C_{\text{exec}}}

\tcbset{
  guidelinebase/.style={
    enhanced,
    breakable,
    colback=blue!4,           
    colframe=blue!50!black,   
    boxrule=0.6pt,            
    arc=2pt,
    left=6pt,right=6pt,top=6pt,bottom=6pt,
  }
}

\newtcolorbox{guidelineboxA}{
  guidelinebase,
  colframe=blue!0,
  boxrule=0pt
}

\newtcolorbox{guidelineboxB}[1]{
  guidelinebase,
  title={#1},
  fonttitle=\bfseries,
  coltitle=black,
  colbacktitle=blue!8,       
  boxed title style={
    boxrule=0pt,
    arc=2pt,
    left=4pt,right=4pt,top=2pt,bottom=2pt
  }
}
\newtcolorbox{guidelineboxC}{
  guidelinebase,
  before skip=6pt,
  after skip=6pt,
  left=5pt,right=5pt,top=5pt,bottom=5pt,
}

\title{When Single-Agent with Skills Replace Multi-Agent Systems and When They Fail}

\author{%
  Xiaoxiao Li\\
  \vspace{0.3em}
  \includegraphics[height=1.5cm]{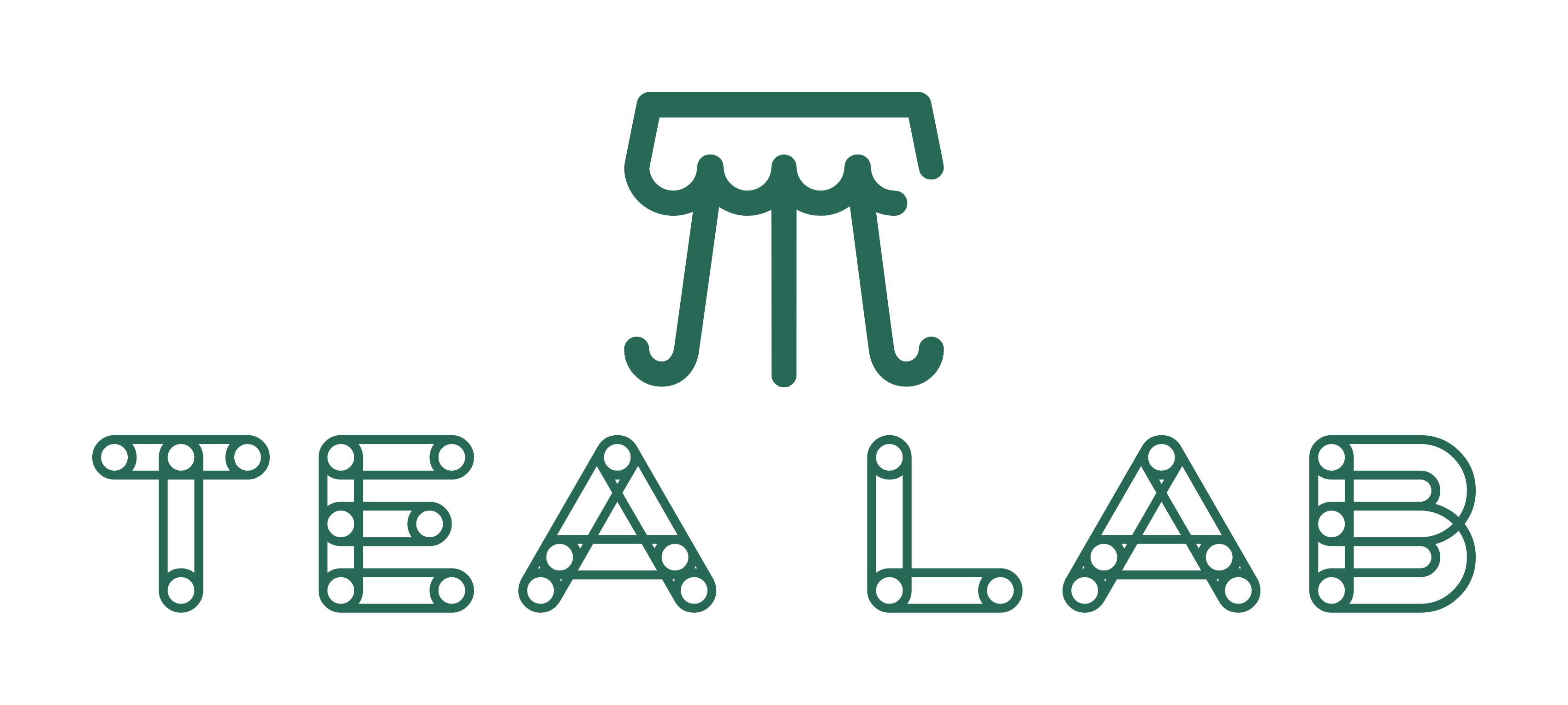}\\
  Trusted and Efficient AI (TEA) Lab\\
  University of British Columbia \textbar{}
  Vector Institute \textbar{} CIFAR AI Chair\\
  \texttt{xiaoxiao.li@ece.ubc.ca}
}

\date{}

\begin{document}

\maketitle

\begin{figure}[t]
    \centering

    \begin{subfigure}[t]{\linewidth}
        \centering
        \includegraphics[width=0.95\linewidth]{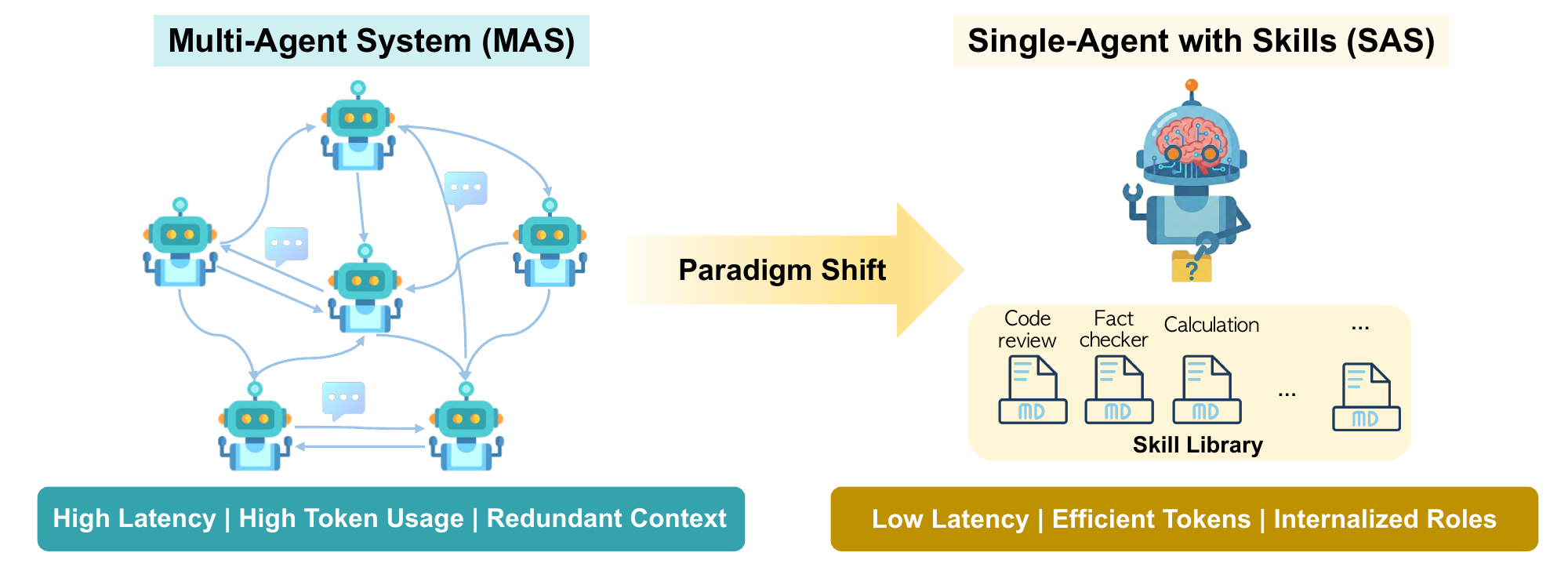}
        \caption{From multi-agent systems to a single agent with skills.}
        \label{fig:skill_scaling_a}
    \end{subfigure}
    \begin{subfigure}[t]{\linewidth}
        \centering
        \includegraphics[width=0.95\linewidth]{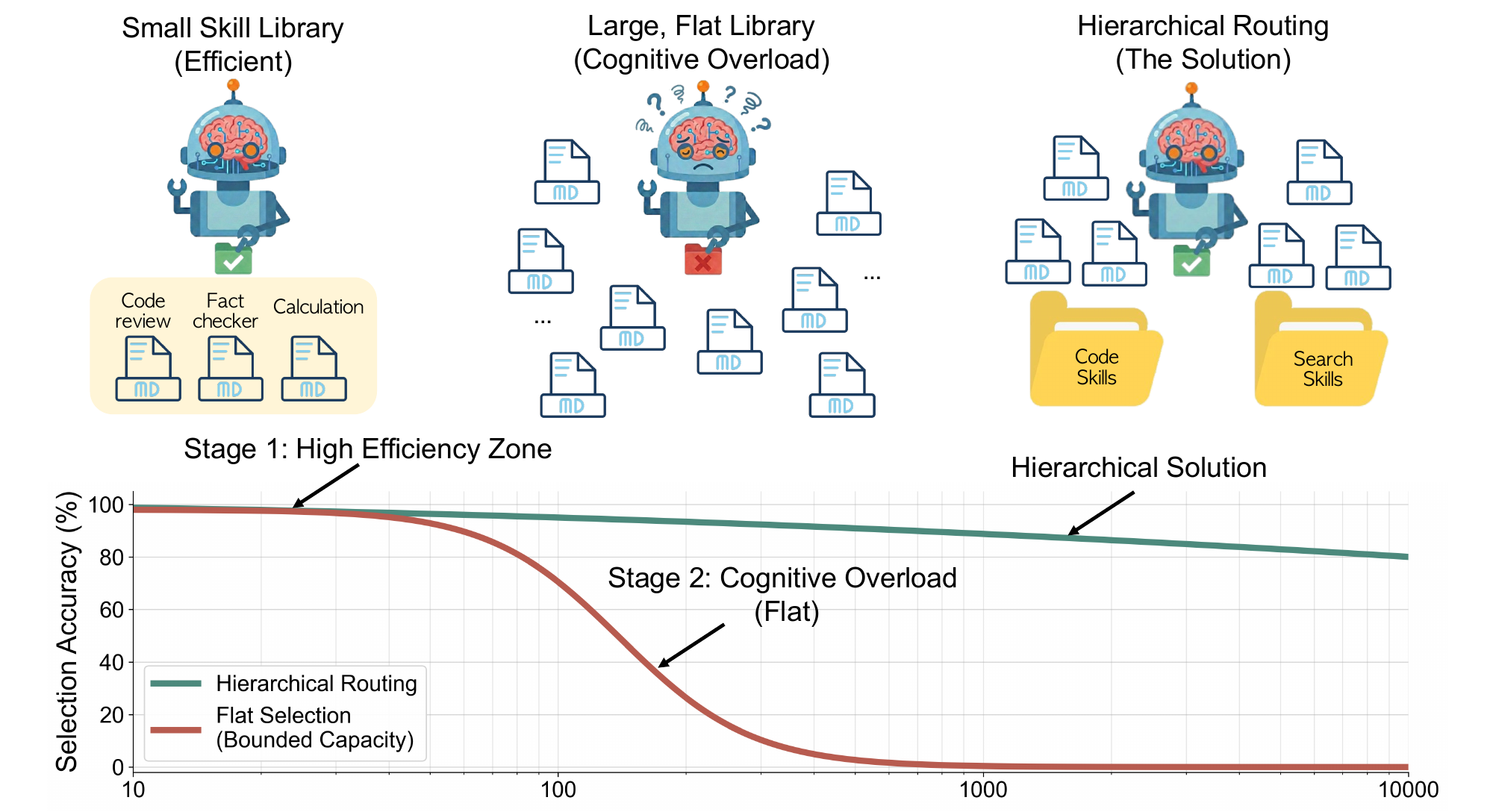}
        \caption{Skill scaling law and its different stages.}
        \label{fig:skill_scaling_b}
    \end{subfigure}

\caption{\textbf{Skill-based agents: efficiency gains and scaling limits.}
(a)~Compiling multi-agent systems into single-agent skill libraries reduces communication overhead, cutting latency and token usage.
(b)~Skill selection accuracy degrades non-linearly as libraries grow, exhibiting a phase transition at a capacity threshold. As skill libraries grow, the increased size and semantic confusability among skills drive this degradation; hierarchical routing restores reliable selection by organizing skills into structured categories.
}

    \label{fig:skill_scaling}
\end{figure}

\begin{abstract}
Multi-agent AI systems have proven effective for complex reasoning. These systems are compounded by specialized agents, which collaborate through explicit communication, but incur substantial computational overhead. A natural question arises: \emph{can we achieve similar modularity benefits with a single agent that selects from a library of skills?} We explore this question by viewing skills as internalized agent behaviors. From this perspective, a multi-agent system can be compiled into an equivalent single-agent system, trading inter-agent communication for skill selection. Our preliminary experiments suggest this approach can substantially reduce token usage and latency while maintaining competitive accuracy on reasoning benchmarks. However, this efficiency raises a deeper question that has received little attention: \emph{how does skill selection scale as libraries grow?} Drawing on principles from cognitive science, we propose that LLM skill selection exhibits bounded capacity analogous to human decision-making.
We investigate the scaling behavior of skill selection and observe a striking pattern.
Rather than degrading gradually, selection accuracy remains stable up to a critical library size, then drops sharply, indicating a phase transition reminiscent of capacity limits in human cognition. Furthermore, we find evidence that semantic confusability among similar skills, rather than library size alone, plays a central role in this degradation.
This perspective suggests that hierarchical organization, which has long helped humans manage complex choices, may similarly benefit AI systems. Our initial results with hierarchical routing support this hypothesis. This work opens new questions about the fundamental limits of semantic-based skill selection in LLMs and offers a cognitive-grounded framework and practical guidelines for designing scalable skill-based agents.
\end{abstract}

\begin{quote}
\textit{``The capacity of the human mind for formulating and solving complex problems is very small compared with the size of the problems whose solution is required for objectively rational behavior in the real world.''}
\begin{flushright}
— \textit{Herbert A. Simon}
\end{flushright}
\end{quote}

\section{Introduction}

Large Language Models (LLMs) are increasingly deployed as general-purpose problem solvers that rely on modular decomposition to handle complex tasks.
Recent progress has shown that \textbf{multi-agent systems} (MAS), where specialized agents collaborate via explicit communication, can substantially improve reasoning performance on challenging benchmarks~\citep{xia2025mmedagent, wu2025agentic, wu2023autogen, chen2024agentverse, guo2024large}.
However, these systems incur \textbf{significant computational overhead} due to repeated context exchange, multi-round coordination, and verbose natural language interactions~\citep{chen2025optima,yang2025autohma,yue2025masrouter}. 
A natural question arises: \textit{can we retain the benefits of modular reasoning while reducing the cost of explicit multi-agent coordination?}

One promising direction is to replace distributed agent coordination with \textbf{tool use}---equipping a single LLM with external APIs that it can invoke as needed~\citep{schick2023toolformer,qin2024toolllm,patil2023gorilla}. 
While effective for atomic operations (calculators, search engines, code interpreters), tools typically lack the rich behavioral specifications needed for complex reasoning subtasks.
In this work, we investigate \textbf{skills} (recently introduced by Anthropic~\citep{anthropic2025skills,anthropic2025skillseng}) as a middle ground: a skill is a \emph{schema-bounded operation} characterized by a semantic descriptor, a well-defined input-output signature, and an execution policy that specifies \emph{how} to perform the operation.
Unlike tools, which are automatically triggered, skills are chosen based on the meaning and content of user requests, thus encapsulating not just \emph{what} to do but \emph{how} to reason, making them suitable for internalizing the specialized roles that would otherwise require separate agents.

Skills offer a compelling alternative to multi-agent coordination. Where an MAS instantiates specialized reasoning as \emph{distributed roles} communicating through natural language, a single-agent system with skills (SAS) internalizes these roles as \emph{selectable actions} within a unified context.
This perspective suggests a \textbf{compilation} view: a MAS can be transformed into an equivalent SAS by encoding each agent's behavior as a skill, eliminating inter-agent communication overhead while preserving functional capability.

We first demonstrate that this compilation is both \emph{faithful} and \emph{efficient}. 
On representative reasoning benchmarks (GSM8K~\citep{cobbe2021gsm8k}, HumanEval~\citep{chen2021humaneval}, HotpotQA~\citep{yang2018hotpotqa}), skill-based single-agent systems achieve similar accuracy of their multi-agent counterparts while reducing token consumption by 54\% and latency by 50\% on average.
These results establish skills as a practical alternative to explicit agent decomposition (at least when skill libraries remain small).

However, expanding the skill repertoire introduces a fundamental challenge. As the number of available skills grows, the model must select the appropriate action from an increasingly large and semantically overlapping set. This raises a question that has received little systematic attention:

\begin{quote}
\emph{How does the size of the skill library affect an LLM's ability to select the correct skill?}
\end{quote}

To answer this question, we study the \textbf{scaling behavior of skill selection} in LLMs.
Drawing on principles from cognitive science, we hypothesize that skill selection exhibits capacity-limited scaling analogous to human decision-making. Our experiments confirm this hypothesis. Across controlled skill libraries ranging from 5 to 200 skills, we find that selection accuracy degrades \emph{non-linearly}, following a phase transition pattern: accuracy remains high when library size is below a critical threshold, then drops sharply beyond this capacity. This degradation is also driven by \emph{semantic confusability} among skills, rather than by library size alone. This finding connects LLM behavior to similarity-based interference in human memory retrieval~\citep{shepard1987toward,anderson1974retrieval}. We further show that \emph{hierarchical routing}: decomposing skill selection into coarse-to-fine decisions can effectively mitigate this degradation when flat selection fails when skill library scales up. This mirrors findings from cognitive science on chunking~\citep{chase1973perception} and menu design~\citep{miller1981depth}.

\paragraph{Contributions.} This work makes three contributions:
\begin{enumerate}[leftmargin=*, nosep]
    \item We demonstrate that skill-based systems can approximate multi-agent performance with significantly lower token usage and latency and formalize the compilation process. 
    
    \item We characterize the non-linear scaling limits of skill selection, identifying a capacity threshold and establishing that semantic confusability, not library size alone, drives degradation.

    \item We show that hierarchical routing mitigates scaling limits, providing cognitive-grounded design principles for scalable skill-based agents.
\end{enumerate}

\noindent Our findings bridge agent skills, multi-agent systems, tool use, and cognitive science, offering both theoretical insights into LLM action selection and practical guidelines for building efficient, scalable agent systems.

\section{Problem Formulation: Multi-Agent Systems to Single-Agent with Skills}
\label{sec:formulation}

To answer the question when \textit{``When Single-Agent with Skills Replace Multi-Agent Systems''}, we formalize the relationship between Multi-Agent Systems (MAS) and Single-Agent with Skills (SAS), establishing the theoretical foundation for studying skill selection scaling. We try to use the simple definition as possible to keep only the notations related to our analysis.

\subsection{Multi-Agent Systems}

\begin{definition}[Agent]
An agent is a tuple $a = (\rho, \phi)$, where:
\begin{itemize}
    \item \textbf{Role description ($\rho$):} a semantic specification of the agent's identity and expertise;
    \item \textbf{Behavioral policy ($\phi$):} instructions governing the agent's reasoning and response generation.
\end{itemize}
\end{definition}

\begin{definition}[Multi-Agent System]
A multi-agent system is a tuple $\mathcal{M} = \langle \mathcal{A}, \mathcal{G}, \Pi \rangle$, where:
\begin{itemize}
    \item $\mathcal{A} = \{a_1, \ldots, a_n\}$ is a set of agents;
    \item $\mathcal{G} = (\mathcal{A}, E)$ is a communication graph specifying permissible mechannels;
    \item $\Pi = (\textsc{Init}, \textsc{Route}, \textsc{Term})$ is a coordination protocol.
\end{itemize}
\end{definition}

\begin{algorithm}[t]
\caption{Multi-Agent System Execution}
\label{alg:mas}
\begin{algorithmic}[1]
\REQUIRE Task $x$, Agent set $\mathcal{A}$, Communication graph $\mathcal{G}$, Protocol $\Pi$
\STATE Initialize history $h \leftarrow x$
\STATE Select initial agent $a^{(0)} \leftarrow \Pi.\textsc{Init}(\mathcal{A}, x)$
\STATE $t \leftarrow 0$
\WHILE{not $\Pi.\textsc{Term}(h)$}
    \STATE $y^{(t)} \leftarrow a^{(t)}.\textsc{Execute}(h)$ \COMMENT{Agent generates response}
    \STATE $h \leftarrow h \oplus y^{(t)}$ \COMMENT{Append to history}
    \STATE $a^{(t+1)} \leftarrow \Pi.\textsc{Route}(h, a^{(t)}, \mathcal{G})$ \COMMENT{Route to next agent}
    \STATE $t \leftarrow t + 1$
\ENDWHILE
\RETURN $h$
\end{algorithmic}
\end{algorithm}

The coordination cost of solving task $x$ over $T$ rounds is:
\begin{equation}
    \Cmas(x) = \sum_{t=1}^{T} |y^{(t)}| + T \cdot c_{\text{sync}},
\end{equation}
where $|y^{(t)}|$ denotes messagelength and $c_{\text{sync}}$ captures synchronization overhead per round.

\subsection{Single-Agent with Skills (SAS)}
\label{sec:single agent}
We now define an alternative paradigm that internalizes multi-agent capabilities within a single model.
\begin{definition}[Skill]
A skill is a tuple $s = (\delta, \pi, \xi)$, where:
\begin{itemize}
    \item \textbf{Skill descriptor ($\delta$):} a semantic description used for skill selection;
    \item \textbf{Execution policy ($\pi$):} instructions specifying how to perform the skill;
    \item \textbf{Execution backend ($\xi \in \mathcal{T} \cup \{\emptyset\}$:)} an external tool $t \in \mathcal{T}$, or $\emptyset$ for internal execution (e.g., a prompt template).
\end{itemize}
\end{definition}

This formulation admits a spectrum of implementations:
\begin{itemize}
    \item \textbf{Internalized skills} ($\xi = \emptyset$): Execution occurs entirely within the model's reasoning process via prompt-based invocation.
    \item \textbf{Externalized skills} ($\xi = t$): The model generates parameters for tool $t$, and execution is delegated externally.
\end{itemize}

\begin{remark}[Separation of Selection and Execution]
The decomposition into $(\delta, \pi, \xi)$ reflects a key insight: \textbf{skill selection} depends primarily on the intent signature $\delta$, while \textbf{skill execution} is governed by $\pi$ and $\xi$. This separation allows us to isolate selection complexity from execution variability, which is a distinction critical to our scaling analysis.
\end{remark}

\begin{definition}[Single-Agent with Skills]
A Single-Agent with Skills (SAS) is a tuple $\mathcal{S} = \langle a, \mathbf{S}, \sigma \rangle$, where:
\begin{itemize}
    \item $a$ is a base language model;
    \item $\mathbf{S} = \{s_1, \ldots, s_k\}$ is a skill library;
    \item $\sigma: \mathcal{H} \times \mathbf{D} \rightarrow \mathbf{S}$ is a \textbf{skill selector} mapping context and skill descriptors $\mathbf{D} = \{\delta_1, \ldots, \delta_k\}$ to a skill.
\end{itemize}
\end{definition}
Note that the selector $\sigma$ operates over skill operations $\mathbf{D}$, not full skill specifications. This reflects the realistic constraint that selection decisions are based on semantic descriptors rather than complete procedural knowledge.
\begin{algorithm}[t]
\caption{Single-Agent with Skills Execution}
\label{alg:SAS}
\begin{algorithmic}[1]
\REQUIRE Task $x$, Skill library $\mathbf{S}$, Selector $\sigma$
\STATE Initialize history $h \leftarrow x$
\STATE $t \leftarrow 0$
\WHILE{not $\textsc{Term}(h)$}
    \STATE $s^{(t)} \leftarrow \sigma(h, \mathbf{D})$ \COMMENT{Select skill based on descriptors}
    \STATE $y^{(t)} \leftarrow \textsc{Execute}(s^{(t)}, h)$ \COMMENT{Execute selected skill}
    \STATE $h \leftarrow h \oplus y^{(t)}$
    \STATE $t \leftarrow t + 1$
\ENDWHILE
\RETURN $h$
\end{algorithmic}
\end{algorithm}

The cost of single-agent using skills over $T$ round is:
\begin{equation}
    \CSAS(x) = \sum_{t=1}^{T'} \Big( \Cselect(\sigma, \mathbf{S}) + \Cexec(s^{(t)}) \Big).
\end{equation}

\begin{remark}
Comparing Algorithms~\ref{alg:mas} and \ref{alg:SAS}, the key structural difference is clear: multi-agent systems \textbf{route between agents} (Line 7 in Alg.~\ref{alg:mas}), while skillful single agents \textbf{select among skills} (Line 4 in Alg.~\ref{alg:SAS}). Compilation transforms the former into the latter.
\end{remark}

\subsection{The Compilation Problem}

We now formalize the transformation from MAS to SAS.

\begin{definition}[Compilation]
A compilation is a mapping $\Phi: \mathcal{M} \rightarrow \mathcal{S}$ that transforms a multi-agent system into a skillful single agent. Given $\mathcal{M} = \langle \mathcal{A},  \mathcal{G}, \Pi \rangle$, the compilation produces $\mathcal{S} = \langle a, \mathbf{S}_\Phi, \sigma_\Phi \rangle$, where each agent's specialized function is distilled into one or more skills.
\end{definition}

We formulate the transition from MAS to SAS as a \textit{Capability Compilation} process. Our compiler $\Phi$ must disentangle the high-level reasoning capabilities embedded in an agent's persona from its execution mechanism. We define the compilation function 
$\Phi: \mathcal{G}_{\text{MAS}} \rightarrow \mathbf{S}$ 
as a composite of three distinct operations: Decomposition, Backend Assignment, and Topology Internalization. The MAS-to-SAS compliation algorithm is presented in Algorithm~\ref{alg:spectrum_compilation} in the Appendix.

\subsubsection{Phase 1: Capability Decomposition}
For each agent $a_i \in \mathcal{A}$ defined by its system prompt $\rho_i$, we first apply a decomposition function $f_{\text{decomp}}$ to extract a set of discrete atomic capabilities $\mathcal{K}_i$:
\begin{equation}
    \mathcal{K}_i = f_{\text{decomp}}(\rho_i) = \{ \kappa_{i,1}, \kappa_{i,2}, \dots \}
\end{equation}
Here, a capability $\kappa$ represents a specific functional unit (e.g.,  ``perform code review'' or ``fetch weather data'') derived from the agent's role description, independent of how it is implemented.

\subsubsection{Phase 2: Backend Assignment}
For each extracted capability $\kappa$, the compiler must determine its execution backend $\xi$. Corresponding to internalized skills and externalized skills defined in Sec~\ref{sec:single agent}. we define the assignment function $f_{\text{backend}}$:

\begin{equation}
    s = (\delta, \pi, \xi) \quad \text{where} \quad \xi = 
    \begin{cases} 
        t \in \mathcal{T} & \text{if } \kappa \text{ requires external grounding (Externalized)} \\
        \emptyset & \text{if } \kappa \text{ is purely cognitive (Internalized)}
    \end{cases}
\end{equation}
Here, the skill descriptor $\delta$ is generated to semanticize $\kappa$ for retrieval (e.g., Skill Name).

\subsubsection{Phase 3: Topology Internalization}

The topology internalization operator transforms the explicit communication edges in $\mathcal{G}_{\text{MAS}}$ into implicit input/output constraints within the skill definitions. Let $\mathcal{N}_{\text{out}}(a_i) = \{a_j \mid (a_i, a_j) \in E\}$ be the set of downstream dependencies for agent $a_i$.

We define the constraint injection function $f_{\text{inject}}$:
\begin{equation}
    \pi_k^{\text{final}} = f_{\text{inject}}(\pi_k^{\text{base}}, \mathcal{N}_{\text{out}}(a_i))
\end{equation}

Mathematically, this can be modeled as appending a constraint set $\mathcal{C}$ to the natural language policy:
\begin{equation}
    \pi_k^{\text{final}} \leftarrow \pi_k^{\text{base}} \oplus \left( \bigcup_{a_j \in \mathcal{N}_{\text{out}}(a_i)} \text{Handover}(a_i \to a_j) \right)
\end{equation}

where $\oplus$ denotes string concatenation or semantic fusion, and $\text{Handover}(\cdot)$ generates instructions ensuring the output format of tool $t_k$ is compatible with the expected input of tools in agent $a_j$.

\subsubsection{Optimization Objective}
The goal of the compilation is to produce a skill library $\mathbf{S}$ that minimizes the \textit{Cognitive Load} $\mathcal{L}(\mathbf{S})$ while maintaining functional equivalence to the original MAS:
\begin{equation}
    \min_{\Phi} \mathcal{L}(\Phi(\mathcal{G}_{\text{MAS}})) \quad \text{s.t.} \quad \text{Perf}(\text{SAS}) \approx \text{Perf}(\text{MAS})
\end{equation}
where $\mathcal{L}$ accounts for both the retrieval complexity (selection noise) and the context consumption of the generated skills.

The final Single-Agent System is thus equipped with the compiled skill library:
\begin{equation}
    \mathbf{S} = \bigcup_{a_i \in \mathcal{A}} \bigcup_{t_k \in \tau_i} \left\{ (t_k, \pi_k^{\text{final}}) \right\}
\end{equation}

\subsection{Properties of MAS-to-SAS Compilation}
\begin{definition}[Behavioral Fidelity]
A compilation $\Phi$ is \textbf{behaviorally faithful} if the output distributions are preserved:
\begin{equation*}
    \forall \tau \in \mathcal{T}: \quad P_{\mathcal{M}}(y \mid \tau) = P_{\Phi(\mathcal{M})}(y \mid \tau).
\end{equation*}
\end{definition}

\begin{definition}[Cost Efficiency]
A compilation $\Phi$ is \textbf{cost-efficient} if there exists a non-trivial subset $\mathcal{T}' \subseteq \mathcal{T}$ such that:
\begin{equation*}
    \forall \tau \in \mathcal{T}': \quad C_{\textsc{SAS}}(\tau) < C_{\textsc{mas}}(\tau).
\end{equation*}
\end{definition}

The central trade-off is clear: compilation eliminates inter-agent communication but introduces a selection bottleneck. Formally, the efficiency condition reduces to:
\begin{equation}
    \underbrace{\sum_{t} C_{\text{select}}(\sigma, |\mathbf{S}|)}_{\text{selection overhead}} < \underbrace{\sum_{t} C_{\text{comm}}(\mathcal{G}, \Pi)}_{\text{communication overhead}}.
    \label{eq:tradeoff}
\end{equation}

\section{Experiments on MAS-to-SAS Compilation}
\label{sec:exp1}
First, we identify the compilable and uncompilable MAS. Then, we present results on the feasibility of compiling MAS to SAS, achieving similar performance at lower cost.

\subsection{Conditions for Compilability}

Not all multi-agent systems can be faithfully compiled into a single skillful agent. We characterize the boundary.

\begin{definition}[Compilability]
A multi-agent system $\mathcal{M}$ is \textbf{compilable} if there exists a compilation $\Phi: \mathcal{M} \rightarrow \mathcal{S}$ satisfying:
\begin{equation*}
    \forall x \in \mathcal{X}: \quad P_{\mathcal{M}}(y \mid x) = P_{\Phi(\mathcal{M})}(y \mid x).
\end{equation*}
\end{definition}

\begin{proposition}[Compilability Conditions]
\label{prop:compilability}
A multi-agent system is compilable if and only if:
\begin{enumerate}[label=\textbf{C\arabic*}., leftmargin=*]
    \item \textbf{Serializable Communication}: $\mathcal{G}$ admits a topological ordering—agent interactions can be sequenced without information loss.
    \item \textbf{Shared History}: Agent outputs depend only on shared history $h$, with no private state.
    \item \textbf{Homogeneous Backbone}: All agents use the same underlying model.
\end{enumerate}
Conversely, compilation fails when agents require \textit{true parallelism} (independent sampling), \textit{private information}, \textit{adversarial objectives}, or \textit{heterogeneous capabilities}.
\end{proposition}

Table~\ref{tab:compilability} summarizes common multi-agent architectures and their compilability.

\begin{table}[htpb]
\centering
\caption{Compilability of common multi-agent architectures.}
\label{tab:compilability}
\renewcommand{\arraystretch}{1.2}
\begin{tabular}{@{}llc@{}}
\toprule
\textbf{Architecture} & \textbf{Structure} & \textbf{Compilable} \\
\midrule
Pipeline & $a_1 \rightarrow a_2 \rightarrow \cdots \rightarrow a_n$ & \cmark \\
Router-Workers & Router $\rightarrow$ \{Workers\} $\rightarrow$ Aggregator & \cmark \\
Iterative Refinement & Writer $\leftrightarrow$ Critic (loop) & \cmark \\
\midrule
Debate / Adversarial & Proponent $\leftrightarrow$ Opponent & \xmark \\
Parallel Sampling & Independent agents, best-of-$n$ & \xmark \\
Private Information & Agents with hidden state & \xmark \\
\bottomrule
\end{tabular}
\end{table}

\begin{remark}[Scope]
This work focuses on compilable systems. We study when compilation is \textit{efficient}---a question orthogonal to when it is \textit{possible}.
\end{remark}


\subsection{Compilation Efficiency Experiments}
\label{sec:compilation_exp}

Having established conditions under which multi-agent systems can be compiled into skillful single agents (Proposition~\ref{prop:compilability}), we now empirically investigate whether such compilation is \emph{efficient}, i.e., whether the compiled SAS achieves comparable performance while reducing computational cost.

\subsubsection{Experimental Setup}

In both internalized and externalized skill cases, the \textit{selection} of which skill to invoke remains an internal cognitive decision. In this studying, we focus on purely internalized skills to isolate the cognitive cost of action selection from external execution noise. We evaluate compilation efficiency on three benchmarks, each matched to a compilable MAS architecture, including GSM8K~\citep{cobbe2021gsm8k} --- grad school math with problems, HumanEval~\citep{chen2021humaneval} --- a python code generation task, and HotpotQA~\citep{yang2018hotpotqa} --- multi-hop question answering. Their corresponding MAS architectures, workflows and their mapping are summarized in Table~\ref{tab:experiment_setup}. The compiled SAS performs the equivalent computation in a \textbf{single API call}, with the model internally managing skill invocation through structured output sections (e.g., \texttt{[DECOMPOSE]}, \texttt{[SOLVE]}, \texttt{[VERIFY]}). 

We evaluate using GPT-4o-mini as the backbone model for all agents and the compiled SAS, ensuring a fair comparison under the homogeneous backbone condition (C3). 
We measure: \textit{Task Accuracy} as correctness of final output (exact match for GSM8K/HotpotQA, syntax validity for HumanEval); \textit{Total Tokens} as the sum of prompt and completion tokens across all API calls; \textit{Latency} as the wall-clock time from task input to final output; and \textit{API Calls} as the number of separate LLM invocations.

\begin{table}[t]
    \centering
    \caption{Benchmark tasks, MAS architectures, and agent-to-skill mappings for compilation experiments. Each MAS is compiled into an equivalent SAS that performs the same computation in a single LLM call.}
    \label{tab:experiment_setup}
    \small
    \resizebox{\linewidth}{!}{
    \begin{tabular}{llllll}
        \toprule
        \textbf{Benchmark} & \textbf{Architecture} & \textbf{MAS Agent} & \textbf{SAS Skill} & \textbf{Description} & \textbf{Calls} \\
        \midrule
        \multirow{3}{*}{GSM8K} 
            & \multirow{3}{*}{Pipeline}
            & Decomposer & \texttt{decompose} & Break problem into steps & \multirow{3}{*}{$3 \to 1$} \\
            & & Solver & \texttt{solve} & Execute calculations & \\
            & & Verifier & \texttt{verify} & Validate solution & \\
        \midrule
        \multirow{3}{*}{HumanEval}
            & \multirow{3}{*}{Iterative}
            & Coder & \texttt{code} & Generate implementation & \multirow{3}{*}{$3 \to 1$} \\
            & & Critic & \texttt{critique} & Review for bugs & \\
            & & Refiner & \texttt{refine} & Fix identified issues & \\
        \midrule
        \multirow{4}{*}{HotpotQA}
            & \multirow{4}{*}{Router-Workers}
            & Router & \texttt{analyze} & Plan information needs & \multirow{4}{*}{$4 \to 1$} \\
            & & Retriever & \texttt{retrieve} & Extract relevant facts & \\
            & & Reasoner & \texttt{reason} & Chain logical steps & \\
            & & Aggregator & \texttt{synthesize} & Produce final answer & \\
        \bottomrule
    \end{tabular}
}
\end{table}

\subsubsection{Results}
\begin{table}[t]
\centering
\caption{Performance and efficiency comparison between MAS and SAS after compilation.
Token and latency reductions are computed as $(\text{MAS} - \text{SAS})/\text{MAS} \times 100\%$.
Positive values indicate SAS is more efficient.}
\label{tab:compilation_results}
\resizebox{\linewidth}{!}{
\begin{tabular}{l|cc|cc|cc|cccc}
\toprule
\multirow{2}{*}{Task} 
& \multicolumn{2}{c|}{Accuracy (\%)} 
& \multicolumn{2}{c|}{Avg. Tokens} 
& \multicolumn{2}{c|}{Avg. Latency (ms)} 
& \multirow{2}{*}{Acc. $\Delta$} 
& \multirow{2}{*}{Token $\downarrow$} 
& \multirow{2}{*}{Latency $\downarrow$} 
& \multirow{2}{*}{API Calls} \\
& MAS & SAS & MAS & SAS & MAS & SAS &  &  &  &  \\
\midrule
GSM8K     
& 94.0 & 92.0 
& 1407 & 616 
& 10565 & 7537 
& $-2.0\%$ 
& $56.2\%$ 
& $28.7\%$ 
& $3 \rightarrow 1$ \\

HumanEval 
& 100.0 & 100.0 
& 1400 & 749 
& 7227 & 2970 
& $0.0\%$ 
& $46.5\%$ 
& $58.9\%$ 
& $3 \rightarrow 1$ \\

HotpotQA  
& 84.0 & 88.0 
& 4359 & 1816 
& 11671 & 4559 
& $+4.0\%$ 
& $58.4\%$ 
& $60.9\%$ 
& $4 \rightarrow 1$ \\

\midrule
Average   
& -- & -- 
& -- & -- 
& -- & -- 
& $\mathbf{+0.7\%}$ 
& $\mathbf{53.7\%}$ 
& $\mathbf{49.5\%}$ 
& -- \\
\bottomrule
\end{tabular}
}
\end{table}

Table~\ref{tab:compilation_results} presents the main results comparing MAS and compiled SAS across all benchmarks. 

\noindent\textbf{Faithful Compilation.} The compiled SAS achieves accuracy within $-2.0\%$ to $+4.0\%$ of the original MAS across all benchmarks, with an average improvement of $+0.7\%$. This confirms that compilation is \emph{faithful} in practice---the SAS preserves (and occasionally improves upon) MAS performance. Notably, on HotpotQA, the SAS outperforms the MAS by $4.0\%$, likely because the unified context enables better information integration across the retrieval and reasoning steps.

\noindent\textbf{Token Efficiency.} Compilation reduces token consumption by 53.7\% on average, with the largest savings on HotpotQA ($58.4\%$) and GSM8K ($56.2\%$). This reduction stems from eliminating redundant context repetition across agent calls---the SAS shares a single context window rather than re-encoding the task description, intermediate results, and instructions for each specialized agent. The effect is most pronounced for HotpotQA, where the multi-hop reasoning pipeline originally required passing retrieved passages between agents multiple times.

\noindent\textbf{Latency Reduction.} End-to-end latency decreases by 49.5\% on average, with the most dramatic improvement on HotpotQA ($60.9\%$) and HumanEval ($58.9\%$). The primary factor is the reduction from 3--4 sequential API calls to a single call, eliminating inter-agent communication overhead and network round-trip delays. GSM8K shows a smaller latency reduction ($28.7\%$) despite significant token savings, suggesting that the mathematical reasoning computation itself (rather than communication) dominates latency for this task.

\noindent\textbf{API Call Reduction.} API calls reduce from 3--4 calls (MAS) to exactly 1 call (SAS). This has direct cost implications, as API pricing typically includes per-call overhead beyond token costs. For HotpotQA, the reduction from 4 calls to 1 represents a $75\%$ decrease in per-request API overhead.

These results demonstrate that MAS-to-SAS compilation has the potential benefits of modular multi-agent architectures (reusable components) combined with the \emph{runtime} efficiency of monolithic single-agent execution. Practitioners can develop and debug agent pipelines using intuitive multi-agent abstractions, then compile to SAS for production deployment without sacrificing accuracy.

The success of compilation rests on three factors: First, agent behaviors are successfully encoded as skills through descriptive prompts. The model can follow structured skill invocation patterns (e.g., ``First decompose, then solve, finally verify''). Second, the sequential structure of agent interactions (C1) maps naturally to a single autoregressive generation. The model maintains coherent state across ``skill invocations'' within its context window. Third, there is no information loss under shared history (C2), as all information available to MAS agents is equally available to the SAS. No private state is lost in compilation.

In this version, we have not explored in depth when complications fail to help, even under Proposition~\ref{prop:compilability}. But we observe that compilation provides smaller benefits when tasks require very long outputs that approach context limits.

\section{The Skill Scaling Hypothesis: A Cognitive Science Perspective}
\label{sec:scaling_hypothesis}

The preceding results establish that MAS $\to$ SAS compilation is both \emph{possible} and \emph{efficient} for compilable architectures. However, this raises a critical follow-up question: \textbf{how does the compiled SAS scale as the skill library grows?}

Equation~\eqref{eq:tradeoff} reveals a critical dependency: the viability of compilation hinges on how selection cost scales with the skill library size $|\mathbf{S}|$. In the experiments above, each SAS operates with 3--4 skills---well below problematic thresholds. When compiling larger multi-agent systems (e.g., with 10+ specialized agents), the resulting skill library may trigger degradation in selection accuracy. This motivates our investigation into the \emph{cognitive scaling} of skill selection.

\subsection{Cognitive Foundations}
\label{sec:cognitive}

Our scaling hypothesis draws on established principles from cognitive psychology and decision science. Four foundational theories are particularly relevant:

\paragraph{F1: Hick's Law and Decision Complexity.}
Hick's Law~\citep{hick1952rate,hyman1953stimulus} establishes that human choice reaction time scales logarithmically with the number of alternatives: $\text{RT} = a + b \cdot \log_2(n+1)$, where $n$ is the number of equally probable options. This reflects a binary subdivision process in decision-making. Critically, \citet{longstreth1988hick} demonstrated that this relationship breaks down beyond approximately 8 choices ($\sim$3 bits), becoming curvilinear as the subdivision strategy fails. We hypothesize that LLM skill selection exhibits analogous capacity-limited behavior, with accuracy (rather than reaction time) as the dependent measure.

\paragraph{F2: Cognitive Load Theory (Working Memory Capacity Limits).}
Miller's seminal ``magical number seven''~\citep{miller1956magical} identified fundamental limits on immediate memory span. \citet{sweller1988cognitive}'s and \citet{sweller2011cognitive}'s Cognitive Load Theory extend this to complex tasks, distinguishing \emph{intrinsic load} (inherent task complexity) from \emph{extraneous load} (unnecessary processing demands). When total cognitive load exceeds working memory capacity, performance degrades sharply---a threshold effect rather than gradual decline. We interpret the skill library size $|\mathbf{S}|$ as imposing intrinsic load on the selection process, with a capacity threshold $\kappa$ analogous to working memory limits.

\paragraph{F3: Similarity-Based Interference.}
Shepard's Universal Law of Generalization~\citep{shepard1987toward} establishes that confusion probability decays exponentially with psychological distance: $g(d) = e^{-d/\lambda}$. Anderson's ACT-R model~\citep{anderson1974retrieval,anderson1983spreading,anderson1999fan} provides a mechanistic account through the \emph{fan effect}: as more facts share a cue, associative strength to each fact decreases, so each receives less activation, leading to slower retrieval and, in some tasks, reduced accuracy. The Generalized Context Model~\citep{nosofsky1986attention} formalizes how classification errors increase with both option count and inter-option similarity. These theories predict that semantically similar skills will interfere during selection, independent of total library size.

\paragraph{F4: Hierarchical Processing and Chunking.}
Chase \& Simon's chunking theory~\citep{chase1973perception} demonstrates that experts manage complexity through hierarchical organization---chess masters perceive board positions as $\sim$7 chunks rather than 32 pieces. Menu design research~\citep{miller1981depth,lee1985optimal} finds optimal breadth of 4--8 items per level, matching working memory capacity. Tversky's Elimination-by-Aspects model~\citep{tversky1972elimination} formalizes how stepwise narrowing of options can make large choice sets more manageable.

\subsection{Problem Formulation}

\begin{problem}[Cognitive Scaling of Skill Selection]
Let $\sigma$ be a skill selector operating on a library $\mathbf{S} = \{s_1, \dots, s_N\}$, where each skill $s_i = (\delta_i, \pi_i, \xi_i)$ comprises a semantic descriptor $\delta$, an execution policy $\pi$, and an execution backend $\xi$. Define the selection accuracy for a task distribution $\mathcal{T}$ as:
\begin{equation*}
\label{eq:accuracy}
    \textsc{Acc}(\sigma, \mathbf{S}) = \mathbb{E}_{\tau \sim \mathcal{T}} \left[ \mathbf{1}\left[ \sigma(\tau, \{ \delta_k \}_{k=1}^N) = s^*_\tau \right] \right],
\end{equation*}
where $s^*_\tau$ is the optimal skill. The scaling problem investigates the degradation of $\textsc{Acc}$ as $|\mathbf{S}| \to \infty$, aiming to decouple the cost of search space expansion from semantic interference.
\end{problem}

\paragraph{Why Study Selection Accuracy?}
One might ask why we focus on skill selection accuracy rather than end-task performance. Task accuracy conflates multiple failure sources: action selection, execution fidelity, and external noise. Changes in selection reliability can be masked at the task level, particularly in error-tolerant settings. In contrast, selection accuracy isolates the intrinsic difficulty of choosing the correct skill from an expanding library. This upstream decision directly governs whether downstream execution is even invoked correctly. While task accuracy answers \emph{whether} a system works, selection accuracy reveals \emph{why} it works---and \emph{when} it will fail.

\subsection{The Scaling Law}

Drawing on the cognitive foundations above, we hypothesize that selection accuracy follows a composite decay law governed by two factors: (1) an effective capacity threshold $\kappa$, analogous to working memory limits, and (2) semantic interference $\mathcal{I}(\mathbf{S})$ among skills:
\begin{equation}
\label{eq:scaling_law}
    \textsc{Acc}(\sigma, \mathbf{S}) \approx \frac{\alpha}{1 + (|\mathbf{S}| / \kappa)^\gamma} - \epsilon \cdot \mathcal{I}(\mathbf{S}),
\end{equation}
where $\alpha \leq 1$ is asymptotic accuracy at small $|\mathbf{S}|$, $\kappa$ is the capacity threshold when accuracy drops to half, $\gamma > 1$ controls the sharpness of the phase transition, and $\epsilon > 0$ represents sensitivity to semantic interference.

This formulation has a clear cognitive interpretation:
\begin{itemize}[leftmargin=*, nosep]
    \item The \textbf{first term} captures Hick's Law--style capacity limits. When $|\mathbf{S}| \ll \kappa$, accuracy remains near $\alpha$; when $|\mathbf{S}| \gg \kappa$, accuracy decays as $O(|\mathbf{S}|^{-\gamma})$. In our model, the exponent $\gamma > 1$ is chosen to capture a super-linear decline in performance once cognitive load exceeds capacity, consistent with cognitive load theory’s assumption of substantial performance degradation under overload~\citep{sweller2011cognitive}.
    
    \item The \textbf{second term} captures Shepard-style similarity interference. Even at fixed $|\mathbf{S}|$, high semantic overlap among skills (large $\mathcal{I}(\mathbf{S})$) degrades accuracy through the ACT-R fan effect---competing skills share retrieval cues, reducing discriminability.
\end{itemize}

The additive structure reflects that capacity limits and similarity interference are \emph{partially independent} failure modes: a small library of highly confusable skills can fail (high $\mathcal{I}$, low $|\mathbf{S}|$), as can a large library of distinct skills (low $\mathcal{I}$, high $|\mathbf{S}|$).

\subsection{Scaling Hypotheses}

The formulation in Eq.~\eqref{eq:scaling_law} implies four testable predictions, each grounded in cognitive theory:

\begin{enumerate}[leftmargin=*]
    \item \textbf{H1: Non-linear Phase Transition.} 
    Selection accuracy exhibits a critical regime governed by capacity threshold $\kappa$. The system maintains high accuracy when $|\mathbf{S}| < \kappa$ but suffers sharp, non-linear degradation once library size exceeds this threshold. This mirrors the breakdown of Hick's Law~\citep{longstreth1988hick} (F1) and the threshold effects in Cognitive Load Theory when working memory is exceeded~\citep{sweller1988cognitive} (F2). The transition is \emph{phase-like} rather than gradual: accuracy is relatively stable until $|\mathbf{S}| \approx \kappa$, then drops precipitously.
    

    \item \textbf{H2: Confusability-Driven Errors.} 
    Selection degradation is driven primarily by \emph{semantic confusability} among skills, not mere library size. Adding semantically similar ``competitor'' skills degrades accuracy more than adding an equivalent number of distinct skills. This prediction follows directly from the interference term $\epsilon \cdot \mathcal{I}(\mathbf{S})$ in Eq.~\eqref{eq:scaling_law} and related to similarity-based interference in cognitive science~\citep{shepard1987toward,anderson1974retrieval,nosofsky1986attention} (F3).
    

    \item \textbf{H3: Instructional Saturation.} 
    Skills encapsulate complex micro-policies $\pi$ that may consume processing bandwidth. We test whether policy complexity affects selection accuracy, specifically, whether verbose policies reduce effective capacity $\kappa$ by increasing extraneous cognitive load~\citep{sweller1988cognitive} (F2).
    

    \item \textbf{H4: Mitigation via Hierarchy.} 
    When flat selection fails ($|\mathbf{S}| > \kappa$), hierarchical organization can restore reliable scaling by ensuring each decision point involves $|\mathbf{S}_{\text{local}}| < \kappa$ options. This transforms an intractable single decision into a sequence of tractable sub-decisions. As described above, chunking theory~\citep{chase1973perception} and Elimination-by-Aspects model~\citep{tversky1972elimination} (F4) establish that hierarchical decomposition to coverts overwhelming choice sets to manageable decisions. 
\end{enumerate}

In the following sections, we empirically test each hypothesis.

\section{Experiments for Scaling-law}
\label{sec:scaling-law experiments}

We design a series of controlled experiments to empirically validate the three key predictions of our skill scaling hypothesis: (1) the existence of a non-linear phase transition in selection accuracy, (2) the effect of instructional complexity on effective capacity, (3) the impact of instructional saturation, and (4) the mitigation potential of hierarchical routing.

\subsection{Experimental Setup}
\label{sec:exp_setup}

\paragraph{Synthetic Skill Library Construction.}
To enable controlled experimentation with precise manipulation of skill library properties, we construct synthetic skill libraries rather than relying on existing benchmarks. This approach allows us to systematically vary library size $|\mathbf{S}|$ while controlling for confounding factors such as semantic similarity distribution and policy complexity.

Our skill generation framework spans 8 domains: \textit{mathematics}, \textit{coding}, \textit{writing}, \textit{analysis}, \textit{translation}, \textit{question-answering}, \textit{formatting}, and \textit{extraction}. Each domain contains 5 subtypes, yielding 40 distinct skill categories. For each category, we define 5 skill templates with varying specificity, producing a pool of 200 unique skill templates. Each skill $s_i = (\delta_i, \pi_i, \xi_i)$ is instantiated with:
\begin{itemize}[leftmargin=*, nosep]
    \item \textbf{Skill descriptor $\delta_i$}: A natural language description following the pattern ``\texttt{[Skill Name]: [Capability Description]}'', e.g., ``\textit{Calculate Sum: Calculate the sum of the given numbers}''.
    \item \textbf{Execution policy $\pi_i$}: Execution instructions with controlled complexity (detailed below).
    \item \textbf{Execution backend $\xi_i$}: Set to $\emptyset$ (internalized) for all experiments unless otherwise specified.
\end{itemize}

\paragraph{Task Generation.}
For each skill library configuration, we generate evaluation tasks by sampling skills uniformly and constructing queries that unambiguously map to the sampled skill. Tasks are generated using domain-specific templates with randomized parameters (e.g., numerical values for math tasks, variable names for coding tasks). Each task $\tau$ is paired with a ground-truth skill label $s^*_\tau$.

\paragraph{Similarity Distribution.}
To control semantic overlap among skills, we define three similarity distributions for library construction:
\begin{itemize}[leftmargin=*, nosep]
    \item \textbf{Low (Diverse)}: Skills are sampled via round-robin across all 8 domains, maximizing semantic distance. Expected intra-library similarity is low.
    \item \textbf{High (Similar)}: Skills are sampled from 2--3 semantically related domains (e.g., \textit{mathematics}, \textit{analysis}, \textit{extraction}). Expected intra-library similarity is high.
    \item \textbf{Mixed}: Skills are sampled uniformly at random across all domains, producing a naturalistic distribution with both related and unrelated skills.
\end{itemize}
Unless otherwise specified, experiments use the mixed distribution to simulate realistic skill library composition.

\paragraph{Policy Complexity.}
To study the effect of execution instruction verbosity, we define three levels of execution policy complexity:
\begin{itemize}[leftmargin=*, nosep]
    \item \textbf{Simple} ($\sim$30 tokens): Single-sentence instructions, e.g., ``\textit{Execute the calculation and return the result.}''
    \item \textbf{Medium} ($\sim$100 tokens): Structured 3--5 step instructions with input validation and formatting guidelines.
    \item \textbf{Complex} ($\sim$300 tokens): Detailed multi-section protocols including error handling, edge cases, output formatting requirements, and quality standards.
\end{itemize}
Unless otherwise specified, experiments use \textbf{simple} policies to isolate the effect of library size from instruction complexity.

\paragraph{Evaluation Protocal.}
We evaluate skill selection performance on GPT-4o-mini and GPT-4o\footnote{Our ongoing work extends the evaluation to additional models.}. The models are queried with temperature $T=0$ to ensure deterministic outputs. The selection prompt presents the skill library as a formatted list and instructs the model to respond with only the skill identifier. For each experimental condition, we measure \textit{selection accuracy} as defined in Eq.~\eqref{eq:accuracy}. Each condition is repeated with 3 random seeds to compute standard errors.

\subsection{H1: Non-linear Phase Transition}
\label{sec:exp_h1}

\paragraph{Experimental Design.}
We evaluate selection accuracy across skill library sizes $|\mathbf{S}| \in \{5, 10, 20, 35, 50, 75, 100, 150, 200\}$, holding similarity distribution (mixed) and policy complexity (simple) constant. This isolates the effect of search space size on selection performance.\footnote{Although increasing $|\mathbf{S}|$ also increases prompt length, our goal is not to disentangle all context-length effects, but to characterize the end-to-end difficulty of selecting among an expanding set of semantically-described actions.} 

\paragraph{Results.}

\begin{figure}[t]
    \centering
    \includegraphics[trim=0 0 0 2cm, clip, width=\linewidth]{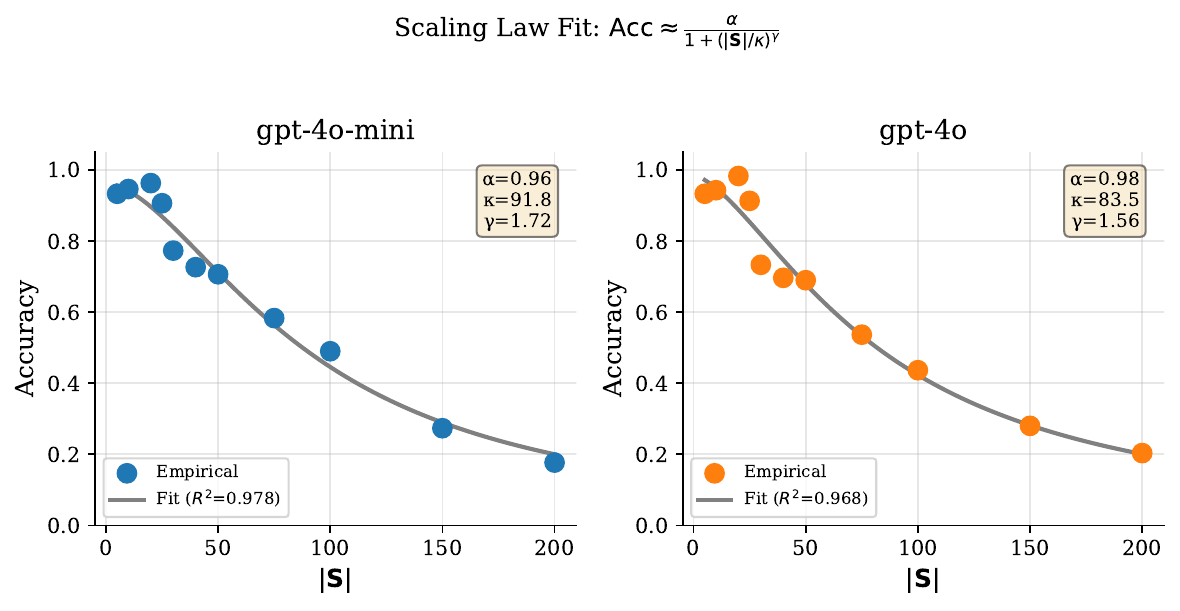}
    \caption{Scaling law fit quality. The proposed functional form $\textsc{Acc} \approx \alpha/(1+(|\mathbf{S}|/\kappa)^\gamma)$ achieves excellent fit ($R^2 > 0.97$) for both models, validating the theoretical model.}
    \label{fig:scaling_law_fit}
\end{figure}

Figure~\ref{fig:scaling_law_fit} presents selection accuracy as a function of skill library size across both evaluated models. We evaluate at $|\mathbf{S}| \in \{5, 10, 20, 25, 30, 40, 50, 75, 100, 150, 200\}$ and fit the proposed scaling law (Eq.~\ref{eq:scaling_law}). Both models exhibit substantial accuracy degradation as library size increases. At small scales ($|\mathbf{S}| \leq 20$), accuracy remains above 90\%, but degrades steadily beyond $|\mathbf{S}| = 30$, falling to approximately 20\% at $|\mathbf{S}| = 200$. We observe a slight increase in accuracy as $|\mathbf{S}|$ grows from 5 to approximately 20. This likely reflects improved task-skill coverage: at very small library sizes, some tasks may lack well-matched skills, while moderate expansion improves coverage before selection overload begins. This non-monotonic behavior at small $|\mathbf{S}|$ does not affect our main findings, which concern the capacity-limited regime at larger library sizes. The decay exponent $\gamma > 1$ (1.72 and 1.56 respectively) indicates super-linear degradation, consistent with phase transition behavior rather than gradual decay. 

The fitted capacity parameter $\kappa$ is 91.8 for GPT-4o-mini and 83.5 for GPT-4o, representing the library size at which accuracy drops to half its maximum ($\alpha/2 \approx 0.48$--$0.49$). Interestingly, GPT-4o exhibits a slightly lower $\kappa$ than GPT-4o-mini, contrary to the intuition that stronger models should have higher capacity. We offer two possible explanations: (1) this difference may reflect fitting variance given limited data points, or (2) model capability and selection capacity may be partially independent; a more capable model is not necessarily better at discriminating among many similar options. Further experiments across diverse model families are needed to clarify this relationship. We note that $\kappa$ should not be interpreted as the ``onset'' of degradation. The inflection point, where accuracy begins to decline most rapidly, occurs earlier. For practical applications, we recommend keeping flat skill libraries below this inflection point, or adopting hierarchical routing for larger libraries. 


\subsection{H2: Confusability-Driven Errors}
\label{sec:exp_h2}

\paragraph{Experimental Design.}
To isolate the effect of confusability from library size, we design a controlled experiment with explicit competitor generation. For each \emph{base skill} (e.g., \texttt{calculate\_sum}: ``Add all numbers 
together''), we generate 0, 1, or 2 \emph{competitor skills} with similar descriptions but \emph{different underlying operations} (e.g., \texttt{compute\_average}: ``Compute the mean of all values'').  Ground truth is ensured by construction: each task is generated to require a specific operation (e.g., ``compute the sum''), and only the base skill performs that operation. Competitors have similar  \emph{descriptions} but would produce incorrect \emph{outputs} if selected. For instance, given input \texttt{[3, 7, 2]} and the task 
``compute the total,'' only \texttt{calculate\_sum} returns the correct answer (\texttt{12}), while \texttt{compute\_average} would return wrong answer (e.g., \texttt{4}).  Thus, the mapping from task to base skill is unambiguous despite surface-level similarity among skill descriptions.

We define three confusability conditions:
\begin{itemize}[leftmargin=*, nosep]
    \item \textbf{No Competitors} ($n_{\text{comp}}=0$): Each skill is semantically distinct. Total library size $|\mathbf{S}| = n_{\text{base}}$.
    \item \textbf{Low Confusability} ($n_{\text{comp}}=1$): Each base skill has one competitor. Total $|\mathbf{S}| = 2 \times n_{\text{base}}$.
    \item \textbf{High Confusability} ($n_{\text{comp}}=2$): Each base skill has two competitors. Total $|\mathbf{S}| = 3 \times n_{\text{base}}$.
\end{itemize}

We vary $n_{\text{base}} \in \{5, 10, 15, 20\}$ and evaluate on GPT-4o-mini and GPT-4o with 3 random seeds per condition.

\paragraph{Results.}

\begin{figure}[t]
    \centering
    \includegraphics[trim=0 0 0 1cm, clip, width=\linewidth]{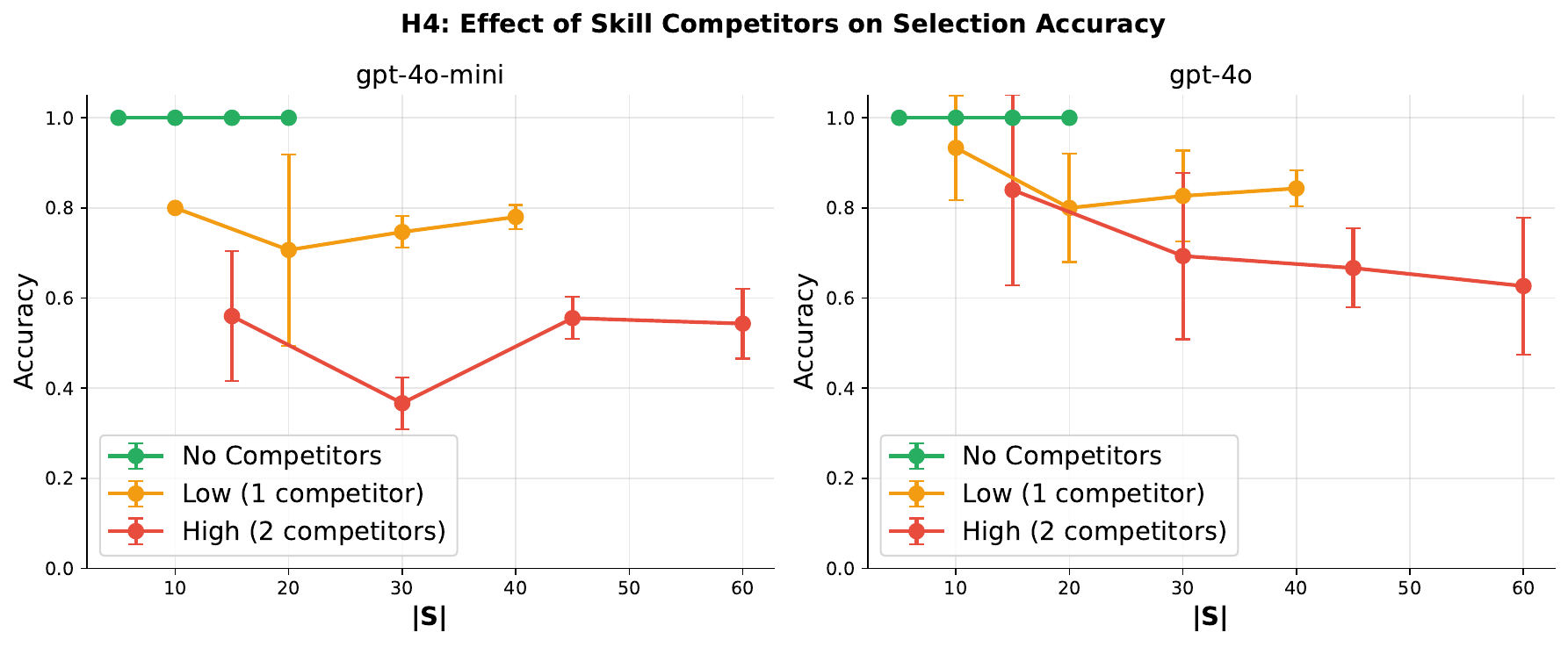}
    \caption{Effect of skill competitors on selection accuracy. Green: no competitors (each skill unique). Orange: 1 competitor per skill. Red: 2 competitors per skill. At fixed total library size, higher confusability leads to lower accuracy, demonstrating that semantic similarity—not library size alone—drives selection errors.}
    \label{fig:h4_confusability}
\end{figure}

Figure~\ref{fig:h4_confusability} presents selection accuracy across confusability conditions. With no competitors, selection accuracy remains at 100\% even at $|\mathbf{S}|=20$. This contrasts sharply with the $\sim$95\% accuracy observed in H1 at the same library size with mixed similarity, confirming that semantic overlap is the primary error source.
    
Adding just one competitor per skill reduces accuracy by 7--30\% (Low condition). Adding two competitors causes 17--63\% degradation (High condition), with GPT-4o-mini showing larger drops than GPT-4o. GPT-4o consistently outperforms GPT-4o-mini under high confusability (e.g., 70\% vs. 37\% at $n_{\text{base}}=10$, $n_{\text{comp}}=2$), suggesting that model capability provides partial mitigation against confusability. The confusability effect is consistent across all tested $n_{\text{base}}$ values, indicating a fundamental challenge rather than a small-scale artifact. At identical $|\mathbf{S}|=20$, replacing unique skills with base-competitor pairs causes a 18--30\% accuracy drop. This demonstrates that \emph{semantic structure} determines selection difficulty.

The results provide an important practical implication for skill library design: skill descriptors should emphasize \emph{unique} characteristics. Avoid generic descriptions that could apply to multiple skills (e.g., ``process data'' $\rightarrow$ ``compute rolling 7-day average'').

H1 and H2 together provide an important understanding of skill selection scaling. H1 establishes that accuracy degrades with library size, while H2 reveals that this degradation is mediated by semantic confusability. The phase transition observed in H1 likely reflects the accumulation of confusable skill pairs as libraries grow from mixed sampling. This explains why the no-competitor condition in H2 maintains 100\% accuracy at $|\mathbf{S}|=20$, a size where H1's mixed-similarity libraries show measurable degradation.

\subsection{H3: Instructional Saturation}
\label{sec:exp_h3}

\paragraph{Experimental Design.}
To test whether execution policy complexity affects effective capacity, we compare selection accuracy across three policy complexity levels (simple, medium, complex) at fixed library sizes $|\mathbf{S}| \in \{10, 20, 50, 100, 150\}$. In this experiment, the full execution policy $\pi_i$ is included in the selection prompt alongside the descriptor $\delta_i$, simulating scenarios where the selector must process detailed execution instructions.

\paragraph{Results.}

\begin{figure}[t]
    \centering
    \includegraphics[trim=0 0 0 1cm, clip, width=\linewidth]{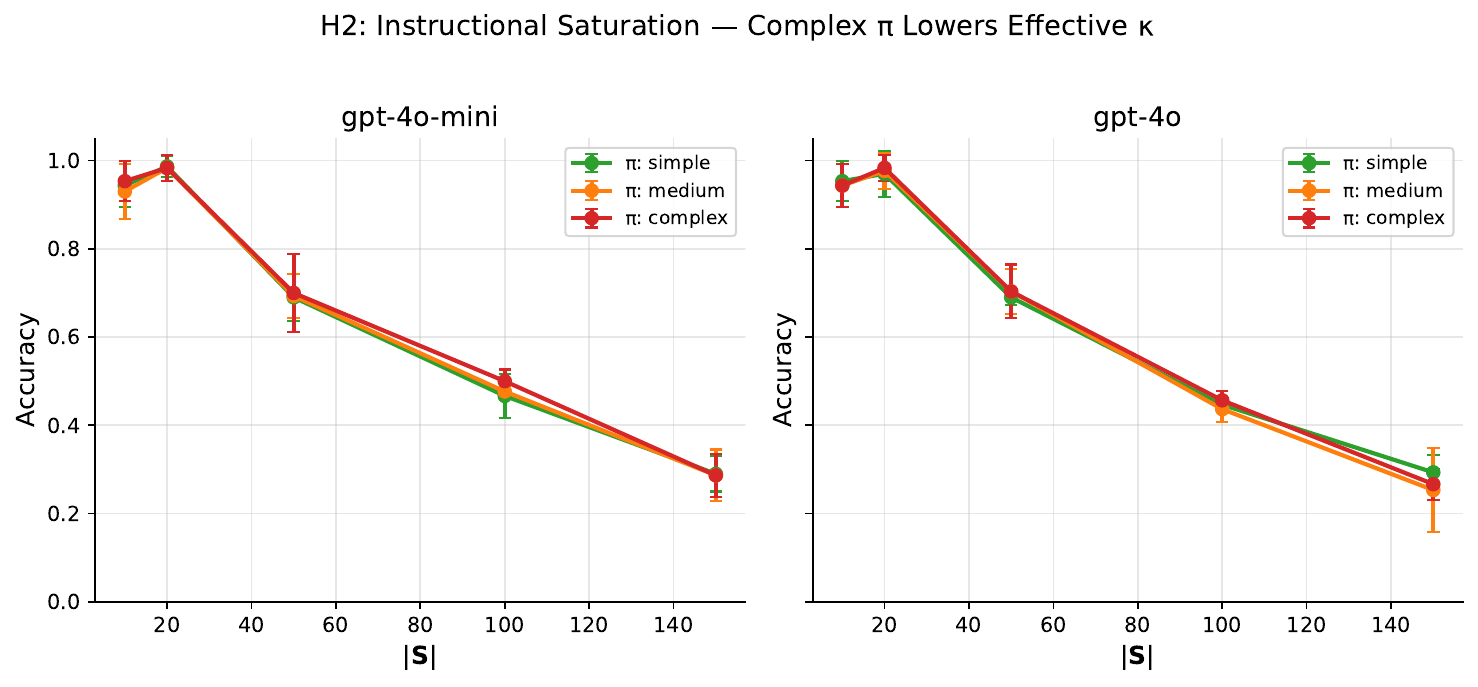}
    \caption{Effect of execution policy complexity on selection accuracy. Each panel shows results for one model. Contrary to expectations, the three complexity levels show largely overlapping performance curves.}
    \label{fig:instructional_saturation}
\end{figure}

Figure~\ref{fig:instructional_saturation} visualizes the interaction between policy complexity and library size for both models. Contrary to our hypothesis, the three policy complexity levels (simple, medium, complex) show largely overlapping accuracy curves for both models. The differences between conditions fall within standard error bounds at most library sizes. All complexity levels follow the same phase transition pattern observed in H1, with accuracy declining sharply beyond $|\mathbf{S}| = 50$. The absence of a complexity effect is consistent across both GPT-4o-mini and GPT-4o, suggesting this is not a model-specific phenomenon.

The results do not support H2. We hypothesized that complex micro-policies would consume ``cognitive bandwidth,'' effectively lowering the capacity threshold $\kappa$. However, the empirical data shows no meaningful difference between simple and complex policies.

Several factors may explain this null result. First, we have to admit that our policy templates, while varying in length, may not have introduced sufficient semantic complexity to stress the selection mechanism. Future work could explore policies with more ambiguous or conflicting instructions. Given the template used, modern transformer architectures may efficiently filter relevant information from long contexts, mitigating the expected cognitive load from complex policies.

\subsection{H4: Hierarchy Mitigation}
\label{sec:exp_h4}

\paragraph{Experimental Design.}
We compare three selection strategies:

\begin{itemize}[leftmargin=*, nosep]
    \item \textbf{Flat Selection}: Direct selection from all $|\mathbf{S}|$ skills (baseline).
    
    \item \textbf{Naive Domain Hierarchy}: Two-stage selection where Stage 1 selects a domain category and Stage 2 selects within that domain.
    
    \item \textbf{Confusability-Aware Hierarchy}: Two-stage selection where semantically similar skills (competitors) are explicitly grouped together. Stage 1 selects among distinct clusters; Stage 2 disambiguates within a small cluster of similar skills.
\end{itemize}

To properly test hierarchy at scale, we extend the library size range beyond the capacity threshold identified in H1. We construct skill libraries with $n_{\text{groups}} \in \{4, 6, 8, 10, 20, 30, 40\}$ groups, each containing 3 semantically similar skills (1 base + 2 competitors), yielding total library sizes $|\mathbf{S}| \in \{12, 18, 24, 30, 60, 90, 120\}$.

Each group represents a distinct functionality (e.g., ``Summation'', ``Averaging'', ``Email Writing''), while skills within a group are near-synonyms (e.g., ``Calculate Sum'', ``Compute Total'', ``Sum Numbers''). Tasks are generated to map to the base skill of each group, ensuring unambiguous ground truth.

\paragraph{Results.}

\begin{figure}[t]
    \centering
    \includegraphics[trim=0 0 0 1cm, clip, width=\linewidth]{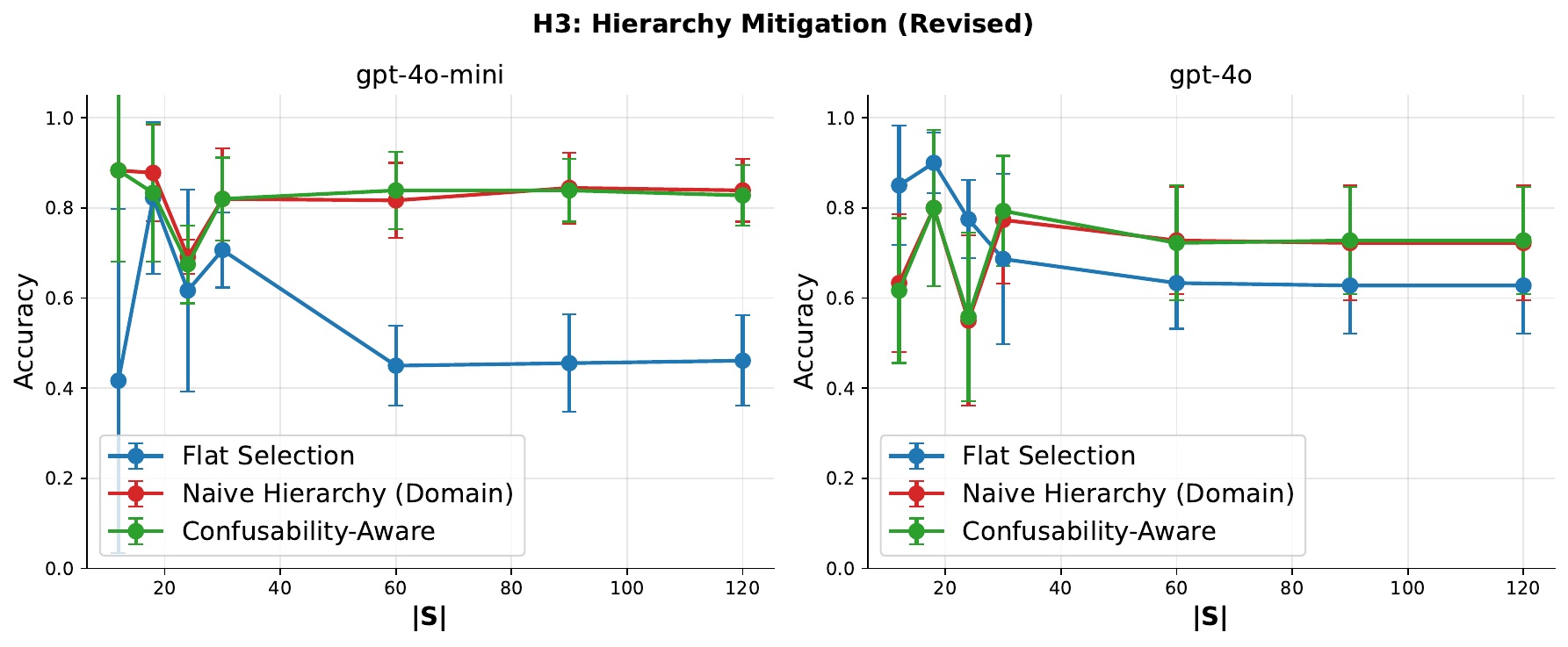}
    \caption{Effect of hierarchical routing on selection accuracy. Blue: flat selection. Red: naive domain hierarchy. Green: confusability-aware hierarchy. At large library sizes ($|\mathbf{S}| \geq 60$), hierarchy maintains $\sim$72--85\% accuracy while flat selection degrades to $\sim$45--63\%.}
    \label{fig:h3_hierarchy}
\end{figure}

Figure~\ref{fig:h3_hierarchy} presents selection accuracy across library sizes for all three methods. When library size exceeds the capacity threshold ($|\mathbf{S}| \geq 60$, approaching $\kappa \approx 90$), hierarchical routing recovers substantial accuracy. For GPT-4o-mini, hierarchy improves accuracy by +37--40\% absolute (from $\sim$45\% to $\sim$83--85\%). For GPT-4o, the improvement is +9--10\% (from $\sim$63\% to $\sim$72\%). Flat selection accuracy drops precipitously as $|\mathbf{S}|$ increases beyond 30, consistent with the phase transition identified in H1. Flat selection at $|\mathbf{S}|=120$ operates far beyond $\kappa$, yielding the observed $\sim$45--63\% accuracy. In contrast, hierarchical methods maintain relatively stable accuracy ($\sim$72--85\%) across all tested sizes up to $|\mathbf{S}|=120$. Both hierarchical methods perform similarly, as our experimental design naturally aligns domain boundaries with confusability clusters. In practice, the key factor is grouping skills such that first-stage routing involves distinct, easily-discriminable categories.GPT-4o shows higher flat accuracy than GPT-4o-mini (63\% vs. 45\% at large $|\mathbf{S}|$), but also derives smaller benefit from hierarchy (+10\% vs. +40\%). Stronger models partially compensate for scaling challenges through better semantic discrimination. Both hierarchical methods perform similarly, as our experimental design naturally aligns domain boundaries with confusability clusters. In practice, the key factor is grouping skills such that first-stage routing involves distinct, easily-discriminable categories.

The results support H3 for library sizes exceeding the capacity threshold. The mechanism can be understood through the lens of H1 and H4.Connecting to H1,  flat selection fails when $|\mathbf{S}| > \kappa$ due to cognitive overload. Hierarchy works by ensuring each selection stage operates within the reliable regime: Stage 1 involves $\sim$10--40 distinct clusters (well below $\kappa$), and Stage 2 involves only 3 skills per cluster. Connecting to H2, confusability-aware grouping ensures that first-stage categories are semantically distinct (low confusability $\rightarrow$ high accuracy), while confusable skills are handled together in a small second-stage pool (high confusability but small $|\mathbf{S}|$ $\rightarrow$ manageable).

\subsection{Practical Guidelines and Limitation Discussion}

Based on our findings, we recommend the following guidelines for skill library design in LLM-based agents:
\begin{guidelineboxB}{Design Guidelines}
Based on our findings, we recommend the following guidelines for skill library design in LLM-based agents:

\begin{enumerate}[leftmargin=*, nosep]
    \item \textbf{Monitor library size}: Track $|\mathbf{S}|$ relative to the model's capacity ($\kappa \approx 50$--$100$ for the tested GPT models). Performance degrades sharply beyond this threshold.
    \item \textbf{Minimize confusability}: Before adding skills, audit semantic overlap. Merge or differentiate skills with similarity above threshold rather than accumulating near-duplicates.
    \item \textbf{Adopt hierarchy at scale}: For large $|\mathbf{S}| \gg \kappa$, implement hierarchical routing with confusability-aware grouping. Each stage should involve $<\kappa$ options.
    \item \textbf{Invest in descriptors}: Since selection relies on descriptors (H2), invest effort in crafting or optimizing distinctive, specific descriptions.
    \item \textbf{Match model to task}: Stronger models show higher $\kappa$ and better confusability resistance. For applications with inherently large or confusable skill libraries, model capability investment yields direct accuracy benefits.
    \item \textbf{Consider alternative architectures}: For large skill libraries, the scaling limitations suggest that multi-agent architectures with specialized routers may outperform monolithic single-agent approaches.
\end{enumerate}
\end{guidelineboxB}

\paragraph{Limitations.}
We acknowledge several limitations. 1. Synthetic data: While enabling controlled experiments, synthetic skill libraries may not capture the full complexity of real-world skill distributions. 2. Selection-only evaluation: We measure selection accuracy but not end-to-end task performance. 3. Limited model coverage: Results are based on two OpenAI models; generalization to other architectures requires further study. 4. Hierarchy design: Designing the best solution to mitigate performance degration associated with scaling is not the focus of this work, so our hierarchical approaches were relatively simple; more sophisticated routing mechanisms may yield different results.

\section{Conclusion and Future Directions}
\label{sec:conclusion}

As LLM-based agents take on increasingly complex tasks, the design of their action spaces becomes a first-order concern. Our work suggests that skill-based modularity offers a promising middle ground between monolithic prompting and expensive multi-agent coordination, but that this approach has inherent scaling limits rooted in the nature of selection itself.  Our preliminary experiments suggest this compilation can yield substantial efficiency gains while preserving task performance---at least when skill libraries remain small. More fundamentally, we investigate a question that has received limited attention: \emph{how does skill selection scale?} Our experiments reveal a non-linear scaling pattern, where selection accuracy remains stable up to a critical threshold before degrading sharply. We find suggestive evidence that semantic confusability among skills also plays a central role in this degradation. Drawing on cognitive psychology, we propose that LLM skill selection may exhibit capacity limits analogous to human decision-making, and that hierarchical organization can help mitigate these limits.

We acknowledge several limitations that may temper the strength of our conclusions. Thus, we acknowledge several \textbf{future directions} worthy of further investigation:

\begin{itemize}[leftmargin=*, nosep]
    \item \textbf{Broader model coverage}: Extending experiments to diverse model families (open-source models, different scales, multimodal models) would clarify whether the capacity thresholds we observe are universal or architecture-specific.
    
    \item \textbf{Real-world skill libraries}: Evaluating on naturally-occurring skill distributions, such as those emerging from software development, long-horizon planning, or scientific workflows, would test the ecological validity of our findings.
    
    \item \textbf{End-to-end evaluation}: Measuring how selection errors propagate to final task outcomes would provide a more complete picture of when skill-based systems are viable alternatives to multi-agent approaches.
    
    \item \textbf{Adaptive routing mechanisms}: Exploring learned or dynamic routing strategies that adapt to task context and skill library structure may offer better scaling than fixed hierarchies.
    
    \item \textbf{Theoretical foundations}: Developing formal models that explain \emph{why} capacity limits emerge and potentially connecting to information-theoretic or mechanistic interpretability perspectives would strengthen the scientific grounding of our observations.
\end{itemize}
We hope the questions and preliminary findings presented here will stimulate further investigation into the cognitive-like constraints that shape what AI agents can effectively do.

\paragraph{Acknowledgment.} This paper represents ongoing research, and we welcome feedback, 
corrections, and suggestions from the community. In accordance with emerging norms for transparency in academic writing, 
we disclose that AI writing assistants were used to support editing and literature organization during the preparation of this manuscript. We have 
carefully reviewed and verified the content. The author thanks Dr. Ruiyang Ge for valuable discussions on cognitive science foundations and for suggesting relevant literature connecting our findings to established theories in human decision-making and is grateful to Yushu Li for providing feedback on early drafts of this manuscript. This work was partially funded by the NSERC Discovery Grant RGPIN-2022-05316, NSERC Alliance Grant ALLRP 602633-24, Tri-Agency Canada IITP, and the Ministry of Science and ICT (No. RS-2024-00445087), CIFAR AI Chair Awards, and Canada Research Chair Fellowship.

\bibliographystyle{plainnat}
\bibliography{reference}

@misc{anthropic2025skills,
  title={Agent Skills Overview},
  author={{Anthropic}},
  year={2025},
  howpublished={\url{https://docs.anthropic.com/en/docs/agents-and-tools/agent-skills/overview}},
  note={Accessed: 2026-01-07}
}

@misc{anthropic2025skillseng,
  title={Equipping Agents for the Real World with Agent Skills},
  author={{Anthropic}},
  year={2025},
  howpublished={\url{https://www.anthropic.com/engineering/equipping-agents-for-the-real-world-with-agent-skills}},
  note={Anthropic Engineering Blog. Accessed: 2026-01-07}
}

@article{cobbe2021gsm8k,
  title={Training Verifiers to Solve Math Word Problems},
  author={Cobbe, Karl and Kosaraju, Vineet and Bavarian, Mohammad and Chen, Mark and Jun, Heewoo and Kaiser, Lukasz and Plappert, Matthias and Tworek, Jerry and Hilton, Jacob and Nakano, Reiichiro and Hesse, Christopher and Schulman, John},
  journal={arXiv preprint arXiv:2110.14168},
  year={2021}
}

@article{chen2021humaneval,
  title={Evaluating Large Language Models Trained on Code},
  author={Chen, Mark and Tworek, Jerry and Jun, Heewoo and Yuan, Qiming and Ponde de Oliveira Pinto, Henrique and Kaplan, Jared and Edwards, Harri and Burda, Yuri and Joseph, Nicholas and Brockman, Greg and others},
  journal={arXiv preprint arXiv:2107.03374},
  year={2021}
}

@inproceedings{yang2018hotpotqa,
  title={{HotpotQA}: A Dataset for Diverse, Explainable Multi-hop Question Answering},
  author={Yang, Zhilin and Qi, Peng and Zhang, Saizheng and Bengio, Yoshua and Cohen, William W. and Salakhutdinov, Ruslan and Manning, Christopher D.},
  booktitle={Proceedings of the 2018 Conference on Empirical Methods in Natural Language Processing},
  pages={2369--2380},
  year={2018}
}

@article{hick1952rate,
  title={On the rate of gain of information},
  author={Hick, William E},
  journal={Quarterly Journal of Experimental Psychology},
  volume={4},
  number={1},
  pages={11--26},
  year={1952}
}

@article{hyman1953stimulus,
  title={Stimulus information as a determinant of reaction time},
  author={Hyman, Ray},
  journal={Journal of Experimental Psychology},
  volume={45},
  number={3},
  pages={188--196},
  year={1953}
}

@article{miller1956magical,
  title={The magical number seven, plus or minus two: Some limits on our capacity for processing information},
  author={Miller, George A},
  journal={Psychological Review},
  volume={63},
  number={2},
  pages={81--97},
  year={1956}
}

@article{sweller1988cognitive,
  title={Cognitive load during problem solving: Effects on learning},
  author={Sweller, John},
  journal={Cognitive Science},
  volume={12},
  number={2},
  pages={257--285},
  year={1988}
}

@book{sweller2011cognitive,
  title={Cognitive Load Theory},
  author={Sweller, John and Ayres, Paul and Kalyuga, Slava},
  year={2011},
  publisher={Springer}
}

@article{shepard1987toward,
  title={Toward a universal law of generalization for psychological science},
  author={Shepard, Roger N},
  journal={Science},
  volume={237},
  number={4820},
  pages={1317--1323},
  year={1987}
}

@article{anderson1974retrieval,
  title={Retrieval of propositional information from long-term memory},
  author={Anderson, John R},
  journal={Cognitive Psychology},
  volume={6},
  number={4},
  pages={451--474},
  year={1974}
}

@article{anderson1983spreading,
  title={A spreading activation theory of memory},
  author={Anderson, John R},
  journal={Journal of verbal learning and verbal behavior},
  volume={22},
  number={3},
  pages={261--295},
  year={1983},
  publisher={Elsevier}
}

@article{anderson1999fan,
  title={The fan effect: New results and new theories.},
  author={Anderson, John R and Reder, Lynne M},
  journal={Journal of Experimental Psychology: General},
  volume={128},
  number={2},
  pages={186},
  year={1999},
  publisher={American Psychological Association}
}

@article{nosofsky1986attention,
  title={Attention, similarity, and the identification--categorization relationship},
  author={Nosofsky, Robert M},
  journal={Journal of Experimental Psychology: General},
  volume={115},
  number={1},
  pages={39--57},
  year={1986}
}

@article{chase1973perception,
  title={Perception in chess},
  author={Chase, William G and Simon, Herbert A},
  journal={Cognitive Psychology},
  volume={4},
  number={1},
  pages={55--81},
  year={1973}
}

@article{tversky1972elimination,
  title={Elimination by aspects: A theory of choice},
  author={Tversky, Amos},
  journal={Psychological Review},
  volume={79},
  number={4},
  pages={281--299},
  year={1972}
}

@article{miller1981depth,
  title={The depth/breadth tradeoff in hierarchical computer menus},
  author={Miller, Dwight P},
  journal={Proceedings of the Human Factors Society},
  volume={25},
  number={1},
  pages={296--300},
  year={1981}
}

@article{lee1985optimal,
  title={Minimizing user search time in menu retrieval systems},
  author={Lee, Eric and MacGregor, James},
  journal={Human Factors},
  volume={27},
  number={2},
  pages={157--162},
  year={1985}
}

@article{longstreth1988hick,
  title={Hick's law: Its limit is 3 bits},
  author={Longstreth, Langdon E},
  journal={Bulletin of the Psychonomic Society},
  volume={26},
  number={1},
  pages={8--10},
  year={1988}
}

@inproceedings{schick2023toolformer,
  title={Toolformer: Language Models Can Teach Themselves to Use Tools},
  author={Schick, Timo and Dwivedi-Yu, Jane and Dess{\`\i}, Roberto and Raileanu, Roberta and Lomeli, Maria and Zettlemoyer, Luke and Cancedda, Nicola and Scialom, Thomas},
  booktitle={Advances in Neural Information Processing Systems (NeurIPS)},
  volume={36},
  pages={68539--68551},
  year={2023}
}

@inproceedings{yao2023react,
  title={ReAct: Synergizing Reasoning and Acting in Language Models},
  author={Yao, Shunyu and Zhao, Jeffrey and Yu, Dian and Du, Nan and Shafran, Izhak and Narasimhan, Karthik and Cao, Yuan},
  booktitle={International Conference on Learning Representations (ICLR)},
  year={2023}
}

@inproceedings{qin2024toolllm,
  title={ToolLLM: Facilitating Large Language Models to Master 16000+ Real-world APIs},
  author={Qin, Yujia and Liang, Shihao and Ye, Yining and Zhu, Kunlun and Yan, Lan and Lu, Yaxi and Lin, Yankai and Cong, Xin and Tang, Xiangru and Qian, Bill and others},
  booktitle={International Conference on Learning Representations (ICLR)},
  year={2024},
  note={Spotlight}
}

@inproceedings{patil2023gorilla,
  title={Gorilla: Large Language Model Connected with Massive APIs},
  author={Patil, Shishir G and Zhang, Tianjun and Wang, Xin and Gonzalez, Joseph E},
  booktitle={Advances in Neural Information Processing Systems (NeurIPS)},
  volume={37},
  year={2024}
}

@inproceedings{shen2023hugginggpt,
  title={HuggingGPT: Solving AI Tasks with ChatGPT and its Friends in Hugging Face},
  author={Shen, Yongliang and Song, Kaitao and Tan, Xu and Li, Dongsheng and Lu, Weiming and Zhuang, Yueting},
  booktitle={Advances in Neural Information Processing Systems (NeurIPS)},
  volume={36},
  pages={38154--38180},
  year={2023}
}

@article{liang2024taskmatrix,
  title={TaskMatrix.AI: Completing Tasks by Connecting Foundation Models with Millions of APIs},
  author={Liang, Yaobo and Wu, Chenfei and Song, Ting and Wu, Wenshan and Xia, Yan and Liu, Yu and Ou, Yang and Lu, Shuai and Ji, Lei and Mao, Shaoguang and others},
  journal={Intelligent Computing},
  volume={3},
  pages={0063},
  year={2024},
  publisher={AAAS}
}

@article{qin2024tool,
  title={Tool Learning with Foundation Models},
  author={Qin, Yujia and Hu, Shengding and Lin, Yankai and Chen, Weize and Ding, Ning and Cui, Ganqu and Zeng, Zheni and Huang, Yufei and Xiao, Chaojun and Han, Chi and others},
  journal={ACM Computing Surveys},
  year={2024},
  publisher={ACM}
}

@article{qu2024tool,
  title={Tool Learning with Large Language Models: A Survey},
  author={Qu, Changle and Dai, Sunhao and Wei, Xiaochi and Cai, Hengyi and Wang, Shuaiqiang and Yin, Dawei and Xu, Jun and Wen, Ji-Rong},
  journal={Frontiers of Computer Science},
  volume={19},
  number={8},
  pages={198343},
  year={2025},
  publisher={Springer}
}

@article{xia2025mmedagent,
  title={MMedAgent-RL: Optimizing Multi-Agent Collaboration for Multimodal Medical Reasoning},
  author={Xia, Peng and Wang, Jinglu and Peng, Yibo and Zeng, Kaide and Wu, Xian and Tang, Xiangru and Zhu, Hongtu and Li, Yun and Liu, Shujie and Lu, Yan and others},
  journal={arXiv preprint arXiv:2506.00555},
  year={2025}
}

@article{wu2025agentic,
  title={Agentic Reasoning: A Streamlined Framework for Enhancing LLM Reasoning with Agentic Tools},
  author={Wu, Junde and Zhu, Jiayuan and Liu, Yuyuan and Xu, Min and Jin, Yueming},
  journal={arXiv preprint arXiv:2502.04644},
  year={2025}
}

@article{yue2025masrouter,
  title={Masrouter: Learning to route llms for multi-agent systems},
  author={Yue, Yanwei and Zhang, Guibin and Liu, Boyang and Wan, Guancheng and Wang, Kun and Cheng, Dawei and Qi, Yiyan},
  journal={arXiv preprint arXiv:2502.11133},
  year={2025}
}

@article{yang2025autohma,
  title={AutoHMA-LLM: Efficient task coordination and execution in heterogeneous multi-agent systems using hybrid large language models},
  author={Yang, Tingting and Feng, Ping and Guo, Qixin and Zhang, Jindi and Ning, Jiahong and Wang, Xinghan and Mao, Zhongyang},
  journal={IEEE Transactions on Cognitive Communications and Networking},
  year={2025},
  publisher={IEEE}
}

@inproceedings{chen2025optima,
  title={Optima: Optimizing effectiveness and efficiency for llm-based multi-agent system},
  author={Chen, Weize and Yuan, Jiarui and Qian, Chen and Yang, Cheng and Liu, Zhiyuan and Sun, Maosong},
  booktitle={Findings of the Association for Computational Linguistics: ACL 2025},
  pages={11534--11557},
  year={2025}
}

@inproceedings{wu2023autogen,
  title={AutoGen: Enabling Next-Gen LLM Applications via Multi-Agent Conversation},
  author={Wu, Qingyun and Bansal, Gagan and Zhang, Jieyu and Wu, Yiran and Li, Beibin and Zhu, Erkang and Jiang, Li and Zhang, Xiaoyun and Zhang, Shaokun and Liu, Jiale and others},
  booktitle={International Conference on Learning Representations (ICLR)},
  year={2024}
}

@inproceedings{hong2024metagpt,
  title={MetaGPT: Meta Programming for A Multi-Agent Collaborative Framework},
  author={Hong, Sirui and Zhuge, Mingchen and Chen, Jonathan and Zheng, Xiawu and Cheng, Yuheng and Zhang, Ceyao and Wang, Jinlin and Wang, Zili and Yau, Steven Ka Shing and Lin, Zijuan and others},
  booktitle={International Conference on Learning Representations (ICLR)},
  year={2024}
}

@inproceedings{li2023camel,
  title={CAMEL: Communicative Agents for ``Mind'' Exploration of Large Language Model Society},
  author={Li, Guohao and Hammoud, Hasan Abed Al Kader and Itani, Hani and Khizbullin, Dmitrii and Ghanem, Bernard},
  booktitle={Advances in Neural Information Processing Systems (NeurIPS)},
  volume={36},
  pages={51991--52008},
  year={2023}
}

@inproceedings{chen2024agentverse,
  title={AgentVerse: Facilitating Multi-Agent Collaboration and Exploring Emergent Behaviors},
  author={Chen, Weize and Su, Yusheng and Zuo, Jingwei and Yang, Cheng and Yuan, Chenfei and Chan, Chi-Min and Yu, Heyang and Lu, Yaxi and Hung, Yi-Hsin and Qian, Chen and others},
  booktitle={International Conference on Learning Representations (ICLR)},
  year={2024}
}

@inproceedings{park2023generative,
  title={Generative Agents: Interactive Simulacra of Human Behavior},
  author={Park, Joon Sung and O'Brien, Joseph C and Cai, Carrie J and Morris, Meredith Ringel and Liang, Percy and Bernstein, Michael S},
  booktitle={Proceedings of the 36th Annual ACM Symposium on User Interface Software and Technology (UIST)},
  pages={1--22},
  year={2023}
}

@article{wang2023voyager,
  title={Voyager: An Open-Ended Embodied Agent with Large Language Models},
  author={Wang, Guanzhi and Xie, Yuqi and Jiang, Yunfan and Mandlekar, Ajay and Xiao, Chaowei and Zhu, Yuke and Fan, Linxi and Anandkumar, Anima},
  journal={Transactions on Machine Learning Research (TMLR)},
  year={2023}
}

@inproceedings{wang2023deps,
  title={Describe, Explain, Plan and Select: Interactive Planning with Large Language Models Enables Open-World Multi-Task Agents},
  author={Wang, Zihao and Cai, Shaofei and Chen, Guanzhou and Liu, Anji and Ma, Xiaojian and Liang, Yitao},
  booktitle={Advances in Neural Information Processing Systems (NeurIPS)},
  volume={36},
  pages={34153--34189},
  year={2023}
}

@inproceedings{guo2024large,
  title={Large Language Model based Multi-Agents: A Survey of Progress and Challenges},
  author={Guo, Taicheng and Chen, Xiuying and Wang, Yaqi and Chang, Ruidi and Pei, Shichao and Chawla, Nitesh V and Wiest, Olaf and Zhang, Xiangliang},
  booktitle={Proceedings of the Thirty-Third International Joint Conference on Artificial Intelligence (IJCAI)},
  pages={8048--8057},
  year={2024}
}

@article{kaplan2020scaling,
  title={Scaling Laws for Neural Language Models},
  author={Kaplan, Jared and McCandlish, Sam and Henighan, Tom and Brown, Tom B and Chess, Benjamin and Child, Rewon and Gray, Scott and Radford, Alec and Wu, Jeffrey and Amodei, Dario},
  journal={arXiv preprint arXiv:2001.08361},
  year={2020}
}

@inproceedings{hoffmann2022training,
  title={Training Compute-Optimal Large Language Models},
  author={Hoffmann, Jordan and Borgeaud, Sebastian and Mensch, Arthur and Buchatskaya, Elena and Cai, Trevor and Rutherford, Eliza and de Las Casas, Diego and Hendricks, Lisa Anne and Welbl, Johannes and Clark, Aidan and others},
  booktitle={Advances in Neural Information Processing Systems (NeurIPS)},
  volume={35},
  pages={30016--30030},
  year={2022}
}

@article{wei2022emergent,
  title={Emergent Abilities of Large Language Models},
  author={Wei, Jason and Tay, Yi and Bommasani, Rishi and Raffel, Colin and Zoph, Barret and Borgeaud, Sebastian and Yogatama, Dani and Bosma, Maarten and Zhou, Denny and Metzler, Donald and others},
  journal={Transactions on Machine Learning Research (TMLR)},
  year={2022}
}

@inproceedings{schaeffer2023emergent,
  title={Are Emergent Abilities of Large Language Models a Mirage?},
  author={Schaeffer, Rylan and Miranda, Brando and Koyejo, Sanmi},
  booktitle={Advances in Neural Information Processing Systems (NeurIPS)},
  volume={36},
  pages={55565--55581},
  year={2023}
}

@inproceedings{michaud2023quantization,
  title={The Quantization Model of Neural Scaling},
  author={Michaud, Eric J and Liu, Ziming and Girit, Uzay and Tegmark, Max},
  booktitle={Advances in Neural Information Processing Systems (NeurIPS)},
  volume={36},
  pages={42540--42569},
  year={2023}
}

@article{wei2022inverse,
  title={Inverse Scaling can become U-shaped},
  author={Wei, Jason and Kim, Najoung and Tay, Yi and Le, Quoc V},
  journal={arXiv preprint arXiv:2211.02011},
  year={2022}
}

@article{mckenzie2023inverse,
  title={Inverse Scaling: When Bigger Isn't Better},
  author={McKenzie, Ian R and Lyzhov, Alexander and Pieler, Michael and Parrish, Alicia and Mueller, Aaron and Prabhu, Ameya and others},
  journal={Transactions on Machine Learning Research (TMLR)},
  year={2023}
}

@article{zhang2024scaling,
  title={When Scaling Meets LLM Finetuning: The Effect of Data, Model and Finetuning Method},
  author={Zhang, Biao and Liu, Zhongtao and Cherry, Colin and Firat, Orhan},
  journal={arXiv preprint arXiv:2402.17193},
  year={2024}
}

@article{kim2025scaling,
  title={Towards a Science of Scaling Agent Systems},
  author={Kim, Yubin and others},
  journal={arXiv preprint arXiv:2512.08296},
  year={2025}
}

\newpage
\appendix
\section{Related Work}
\label{sec:related_work}

In addition to the related work discussion on cognitive science in Sec~\ref{sec:cognitive}, we also relate our work to the intersection of three research areas: tool use in LLMs, multi-agent systems, and scaling laws. Our contribution is distinguished by its focus on the \emph{scaling behavior of skill selection}---a problem that has received limited systematic study despite its centrality to agent system design.

\subsection{Tool Use in LLMs}

Recent work has dramatically expanded LLMs' capabilities by augmenting them with external tools. \citet{schick2023toolformer} introduced Toolformer, demonstrating that LLMs can learn to use tools (calculators, search engines, translation systems) in a self-supervised manner. ReAct~\citep{yao2023react} established the influential paradigm of interleaving reasoning traces with tool-use actions, enabling dynamic decisions about when to invoke external capabilities versus rely on internal knowledge.

As tool ecosystems have grown, the challenge of tool selection has become increasingly prominent. ToolLLM~\citep{qin2024toolllm} scaled to over 16,000 real-world APIs, introducing a neural retriever for tool selection and depth-first search for multi-step planning. Gorilla~\citep{patil2023gorilla} demonstrated retriever-aware training for API selection, while HuggingGPT~\citep{shen2023hugginggpt} and TaskMatrix.AI~\citep{liang2024taskmatrix} proposed architectures where LLMs orchestrate specialized models based on task descriptions. Comprehensive surveys~\citep{qin2024tool,qu2024tool} have formalized tool learning into stages: planning, selection, calling, and response generation.

Our work differs from this literature in two key respects. First, we distinguish \emph{skills} from \emph{tools}: while tools are typically atomic external APIs with minimal descriptive overhead, skills are \emph{internalized capabilities} comprising rich semantic descriptors and execution policies. This distinction matters because skills impose greater cognitive load during selection. Second, rather than proposing new selection mechanisms, we characterize the fundamental scaling limits of selection accuracy by establishing capacity thresholds ($\kappa$) and identifying confusability as the primary driver of degradation.

\subsection{Multi-Agent LLM Systems}

Multi-agent systems (MAS) have emerged as a prominent paradigm for complex task solving. AutoGen~\citep{wu2023autogen} introduced flexible multi-agent conversation frameworks with customizable agent interactions. MetaGPT~\citep{hong2024metagpt} encodes Standardized Operating Procedures (SOPs) into agent workflows, demonstrating that structured role assignment enables effective coordination. CAMEL~\citep{li2023camel} pioneered role-playing frameworks for autonomous agent cooperation, while AgentVerse~\citep{chen2024agentverse} explored dynamic agent group composition with expert recruitment mechanisms.

A parallel line of work has focused on skill acquisition and management in agents. Voyager~\citep{wang2023voyager} introduced an ever-growing skill library for embodied agents, where skills are indexed by embeddings and retrieved compositionally. Generative Agents~\citep{park2023generative} established foundational architectures with memory, reflection, and planning modules. DEPS~\citep{wang2023deps} addressed skill selection through interactive planning with sub-goal ranking.

Our work provides a theoretical bridge between MAS and single-agent skill-based systems. We show that certain types of MAS can be compiled into equivalent single-agent systems with skill libraries, and characterize when this compilation is beneficial versus when the resulting skill library exceeds cognitive capacity. This perspective reveals that hierarchical multi-agent organization is not merely an architectural choice but a necessary mitigation when skill libraries grow beyond the capacity threshold $\kappa$.

\subsection{Scaling Laws and Emergent Abilities}

Neural scaling laws~\citep{kaplan2020scaling,hoffmann2022training} have established power-law relationships between model size, data, compute, and loss. These findings have been extended to understand emergent abilities---capabilities that appear suddenly at specific scale thresholds rather than improving gradually~\citep{wei2022emergent}. The nature of these phase transitions remains debated. \citet{schaeffer2023emergent} argued that apparent emergent abilities may be artifacts of nonlinear metrics rather than fundamental behavioral changes. \citet{michaud2023quantization} proposed that network capabilities are ``quantized'' into discrete skills learned in order of decreasing frequency, with smooth aggregate scaling masking discrete phase transitions in individual capabilities. Recent work has documented U-shaped~\citep{wei2022inverse} and inverted-U scaling patterns~\citep{mckenzie2023inverse}, where performance can decrease before improving at larger scales.

Scaling laws for downstream tasks and agent systems have received growing attention. \citet{zhang2024scaling} studied how fine-tuning data scales with model size, while \citet{kim2025scaling} derived quantitative principles for agent system scaling, examining trade-offs between agent quantity, coordination structure, and task properties.

Unlike prior work on emergent abilities that focuses on model scale, we study scaling with respect to \emph{action space size}---the number of skills an agent must select among. We show that this scaling is mediated by semantic confusability, connecting LLM behavior to established principles from cognitive psychology~\citep{hick1952rate,shepard1987toward,sweller1988cognitive}.


\section{MAS-to-SAS Compilaton Algorithm}

\begin{algorithm}[h]
\caption{MAS-to-SAS Compilation}
\label{alg:spectrum_compilation}
\begin{algorithmic}[1]
\REQUIRE 
    Multi-Agent Graph $\mathcal{G} = (\mathcal{A}, E)$, 
    Compiler Model $\mathcal{M}_{\text{LLM}}$
\ENSURE 
    Unified Skill Library $\mathbf{S}$

\STATE Initialize $\mathbf{S} \leftarrow \emptyset$

\COMMENT{\textbf{Phase 1: Capability Extraction \& Classification}}
\FORALL{Agent $a_i \in \mathcal{A}$}
    \STATE Let $\rho_i$ be the system prompt (role definition)
    \STATE Let $\tau_i$ be the set of explicitly assigned tools
    
    \STATE \textbf{Analyze Role:} Decompose $\rho_i$ into a set of discrete capabilities $\mathcal{C}_i = \{c_1, c_2, \dots\}$.
    
    \FORALL{Capability $c_k \in \mathcal{C}_i$}
        \STATE \textbf{Determine Backend:}
        \IF{$c_k$ aligns with a tool $t \in \tau_i$}
            \STATE Set $\xi_k \leftarrow t$ \COMMENT{Externalized Skill}
            \STATE Generate $\pi_k$: ``How to use $t$ to achieve $c_k$ under role $\rho_i$''
        \ELSE
            \STATE Set $\xi_k \leftarrow \emptyset$ \COMMENT{Internalized Skill}
            \STATE Generate $\pi_k$: ``Reasoning steps to perform $c_k$''
        \ENDIF
        
        \STATE Generate Skill Descriptor $\delta_k$ (Name + Description)
        \STATE Store $\hat{s}_k = (\delta_k, \pi_k, \xi_k, \text{owner}=a_i)$
    \ENDFOR
\ENDFOR

\COMMENT{\textbf{Phase 2: Topology Internalization (Constraints)}}
\FORALL{Skill $\hat{s}_k \in \text{extracted skills}$}
    \STATE Let $a_{\text{src}} = \hat{s}_k.\text{owner}$
    \STATE Identify downstream agents: $\mathcal{N}_{\text{out}} = \{a_j \mid (a_{\text{src}}, a_j) \in E\}$
    
    \IF{$\mathcal{N}_{\text{out}} \neq \emptyset$}
        \STATE \textbf{Generate Handoff Constraint:}
        \STATE ``Output must be consumable by skills: $[\text{skills of } \mathcal{N}_{\text{out}}]$''
        \STATE $\pi_k \leftarrow \pi_k \oplus \text{Constraint}$
    \ENDIF
    
    \STATE Add $(\delta_k, \pi_k, \xi_k)$ to $\mathbf{S}$
\ENDFOR

\RETURN $\mathbf{S}$
\end{algorithmic}
\end{algorithm}

\section{Experimental Details}
\label{app:experimental_details}

\subsection{Skill Library Examples}
\label{app:skill_examples}

Table~\ref{tab:skill_examples} presents example skills from each domain in our synthetic skill library. Each skill consists of a unique identifier, a natural language descriptor, and an execution policy.

\begin{table}[h]
\centering
\caption{Example skills from different domains in our synthetic skill library.}
\label{tab:skill_examples}
\small
\begin{tabular}{llp{7cm}}
\toprule
\textbf{Domain} & \textbf{Skill Name} & \textbf{Descriptor} \\
\midrule
Mathematics & Calculate Sum & Add all numbers together and return the total sum. \\
 & Calculate Average & Compute the arithmetic mean of the given numbers. \\
 & Calculate Percentage & Compute what percentage one number is of another. \\
\addlinespace
\midrule
Coding & Write Python Function & Write a Python function that implements the specified functionality. \\
 & Debug Code & Find and fix bugs in the provided code snippet. \\
\addlinespace
\midrule
Writing & Write Email & Compose a professional email based on the given requirements. \\
 & Write Summary & Create a concise summary of the provided text. \\
\addlinespace
\midrule
Analysis & Analyze Sentiment & Determine the emotional tone (positive/negative/neutral) of the text. \\
 & Analyze Trend & Identify patterns and trends in the provided data. \\
\addlinespace
\midrule
Extraction & Extract Names & Identify and extract all person names mentioned in the text. \\
 & Extract Dates & Identify and extract all dates mentioned in the text. \\
\addlinespace
\midrule
Translation & Translate to Spanish & Translate the given English text into Spanish. \\
 & Translate to French & Translate the given English text into French. \\
\addlinespace
\midrule
QA & Answer Question & Provide a direct answer to the given question. \\
 & Explain Concept & Explain the given concept in clear, simple terms. \\
\addlinespace
\midrule
Formatting & Convert to JSON & Convert the given data into JSON format. \\
 & Format as Markdown & Format the given content as Markdown. \\
\bottomrule
\end{tabular}
\end{table}

\subsection{Task Examples}
\label{app:task_examples}

Table~\ref{tab:task_examples} shows example tasks with their ground-truth skill mappings. Tasks are generated using domain-specific templates with randomized parameters.

\begin{table}[h]
\centering
\caption{Example tasks with ground-truth skill mappings.}
\label{tab:task_examples}
\small
\begin{tabular}{lp{8cm}l}
\toprule
\textbf{Domain} & \textbf{Task Query} & \textbf{Ground Truth} \\
\midrule
Mathematics & What is the sum of 23, 45, and 67? & Calculate Sum \\
Mathematics & What is the average of 10, 20, 30, 40, 50? & Calculate Average \\
Mathematics & What is 25\% of 200? & Calculate Percentage \\
\addlinespace
Coding & Write a Python function to reverse a string. & Write Python Function \\
Coding & Debug this code: \texttt{def add(a,b): return a-b} & Debug Code \\
\addlinespace
Writing & Write an email to my manager requesting a meeting. & Write Email \\
Writing & Summarize the following article: [text] & Write Summary \\
\addlinespace
Analysis & What is the sentiment of: ``I love this product!'' & Analyze Sentiment \\
\addlinespace
Extraction & Extract names from: ``John met Mary at the park.'' & Extract Names \\
Extraction & Extract dates from: ``Meeting on Jan 15, 2024.'' & Extract Dates \\
\addlinespace
Translation & Translate to Spanish: ``Hello, how are you?'' & Translate to Spanish \\
\bottomrule
\end{tabular}
\end{table}

\subsection{Competitor Skill Examples}
\label{app:competitor_examples}

Table~\ref{tab:competitor_examples} illustrates the base skill and competitor skill design used in H2 experiments. Competitors have near-identical functionality but different surface forms, creating semantic confusability.

\begin{table}[h]
\centering
\caption{Examples of base skills and their semantic competitors (H2). Competitors have similar functionality but different phrasing.}
\label{tab:competitor_examples}
\small
\begin{tabular}{lp{10cm}}
\toprule
\textbf{Role} & \textbf{Skill Descriptor} \\
\midrule
\multicolumn{2}{l}{\textit{Skill Group: Summation}} \\
\addlinespace
Base & Calculate Sum: Add all numbers together and return the total. \\
Competitor 1 & Compute Total: Compute the total by adding all values together. \\
Competitor 2 & Sum Numbers: Sum up all the given numbers. \\
\midrule
\multicolumn{2}{l}{\textit{Skill Group: Email Writing}} \\
\addlinespace
Base & Write Email: Compose a professional email. \\
Competitor 1 & Compose Email: Compose an email message. \\
Competitor 2 & Draft Email: Draft an email for the given purpose. \\
\midrule
\multicolumn{2}{l}{\textit{Skill Group: Sentiment Analysis}} \\
\addlinespace
Base & Analyze Sentiment: Determine the sentiment of the text. \\
Competitor 1 & Sentiment Detector: Detect the emotional tone. \\
Competitor 2 & Evaluate Sentiment: Evaluate text sentiment. \\
\midrule
\multicolumn{2}{l}{\textit{Skill Group: Name Extraction}} \\
\addlinespace
Base & Extract Names: Extract person names from text. \\
Competitor 1 & Find Names: Find all names in the text. \\
Competitor 2 & Name Extractor: Extract names from content. \\
\bottomrule
\end{tabular}
\end{table}

\subsection{Policy Complexity Examples}
\label{app:policy_examples}

Figure~\ref{fig:policy_examples} illustrates the three levels of policy complexity used in H3 experiments, using the ``Calculate Sum'' skill as an example.

\begin{figure}[h]
\centering
\begin{tcolorbox}[title=Simple Policy ($\sim$30 tokens), colback=green!5, width=0.95\textwidth]
\small\ttfamily
Execute the addition and return the result.
\end{tcolorbox}

\vspace{0.3cm}

\begin{tcolorbox}[title=Medium Policy ($\sim$100 tokens), colback=yellow!5, width=0.95\textwidth]
\small\ttfamily
You are a mathematical computation assistant.\\
1. Parse all numerical values from the input\\
2. Validate that all values are valid numbers\\
3. Compute the sum of all values\\
4. Return the result as a single number
\end{tcolorbox}

\vspace{0.3cm}

\begin{tcolorbox}[title=Complex Policy ($\sim$300 tokens), colback=red!5, width=0.95\textwidth]
\small\ttfamily
You are an expert mathematical computation agent.\\[0.2cm]
\textbf{Input Processing}\\
1. Parse all numerical values from the input string\\
2. Handle both integers and floating-point numbers\\
3. Validate that all extracted values are valid numbers\\[0.2cm]
\textbf{Computation}\\
4. Initialize accumulator to zero\\
5. Iterate through all validated numbers\\
6. Add each number to the accumulator\\[0.2cm]
\textbf{Output Requirements}\\
7. Return the final sum as a formatted number\\
8. Use appropriate decimal precision (2 decimal places)\\
9. Handle edge cases: empty input returns 0\\[0.2cm]
\textbf{Error Handling}\\
- If non-numeric values are found, report an error\\
- If input is empty, return 0 with a note
\end{tcolorbox}
\caption{Examples of the three policy complexity levels for the ``Calculate Sum'' skill.}
\label{fig:policy_examples}
\end{figure}

\subsection{Prompt Templates}
\label{app:prompt_templates}

\paragraph{Flat Selection Prompt.}
Figure~\ref{fig:flat_prompt} shows the prompt template used for flat skill selection in all experiments.

\begin{figure}[h]
\centering
\begin{tcolorbox}[title=Flat Selection Prompt, colback=blue!5, width=0.95\textwidth]
\small\ttfamily
Select the most appropriate skill for the given task.\\[0.2cm]
Task: \{query\}\\[0.2cm]
Available Skills:\\
- skill\_001: Calculate Sum: Add all numbers together and return the total.\\
- skill\_002: Calculate Average: Compute the arithmetic mean of the given numbers.\\
- skill\_003: Write Email: Compose a professional email.\\
- skill\_004: Extract Names: Identify and extract all person names from text.\\
- skill\_005: Translate to Spanish: Translate English text into Spanish.\\[0.2cm]
Respond with ONLY the skill ID (e.g., skill\_001). Do not include any explanation.
\end{tcolorbox}
\caption{Example flat selection prompt with 5 skills.}
\label{fig:flat_prompt}
\end{figure}

\paragraph{Hierarchical Selection Prompts.}
Figure~\ref{fig:hier_prompt} shows the two-stage prompts used for hierarchical selection in H4.

\begin{figure}[h]
\centering
\begin{tcolorbox}[title=Stage 1: Category Selection, colback=green!5, width=0.95\textwidth]
\small\ttfamily
Select the most appropriate skill category for this task.\\[0.2cm]
Task: What is the sum of 23, 45, and 67?\\[0.2cm]
Available Categories:\\
- Summation: Adding numbers together\\
- Averaging: Computing mean values\\
- Email Writing: Composing emails\\
- Sentiment Analysis: Analyzing emotional tone\\
- Name Extraction: Extracting names from text\\[0.2cm]
Respond with ONLY the category name.
\end{tcolorbox}

\vspace{0.3cm}

\begin{tcolorbox}[title=Stage 2: Skill Selection within Category, colback=orange!5, width=0.95\textwidth]
\small\ttfamily
Select the most appropriate skill for this task.\\[0.2cm]
Task: What is the sum of 23, 45, and 67?\\[0.2cm]
Available Skills in "Summation" category:\\
- skill\_001: Calculate Sum: Add all numbers together and return the total.\\
- skill\_002: Compute Total: Compute the total by adding all values.\\
- skill\_003: Sum Numbers: Sum up all the given numbers.\\[0.2cm]
Respond with ONLY the skill ID.
\end{tcolorbox}
\caption{Two-stage hierarchical selection prompts used in H4.}
\label{fig:hier_prompt}
\end{figure}

\end{document}